\pdfoutput=1

\documentclass[11pt]{article}

\usepackage[]{acl}

\usepackage{times}
\usepackage{latexsym}
\usepackage{xpinyin}
\usepackage{CJKutf8}
\usepackage{subcaption}
\usepackage{url}
\usepackage{booktabs}
\usepackage{multirow}
\usepackage[T1]{fontenc}

\usepackage[utf8]{inputenc}
\usepackage{booktabs}
\usepackage{microtype}

\usepackage{inconsolata}

\usepackage{graphicx}
\usepackage{amsmath}

\title{\includegraphics[height=1em]{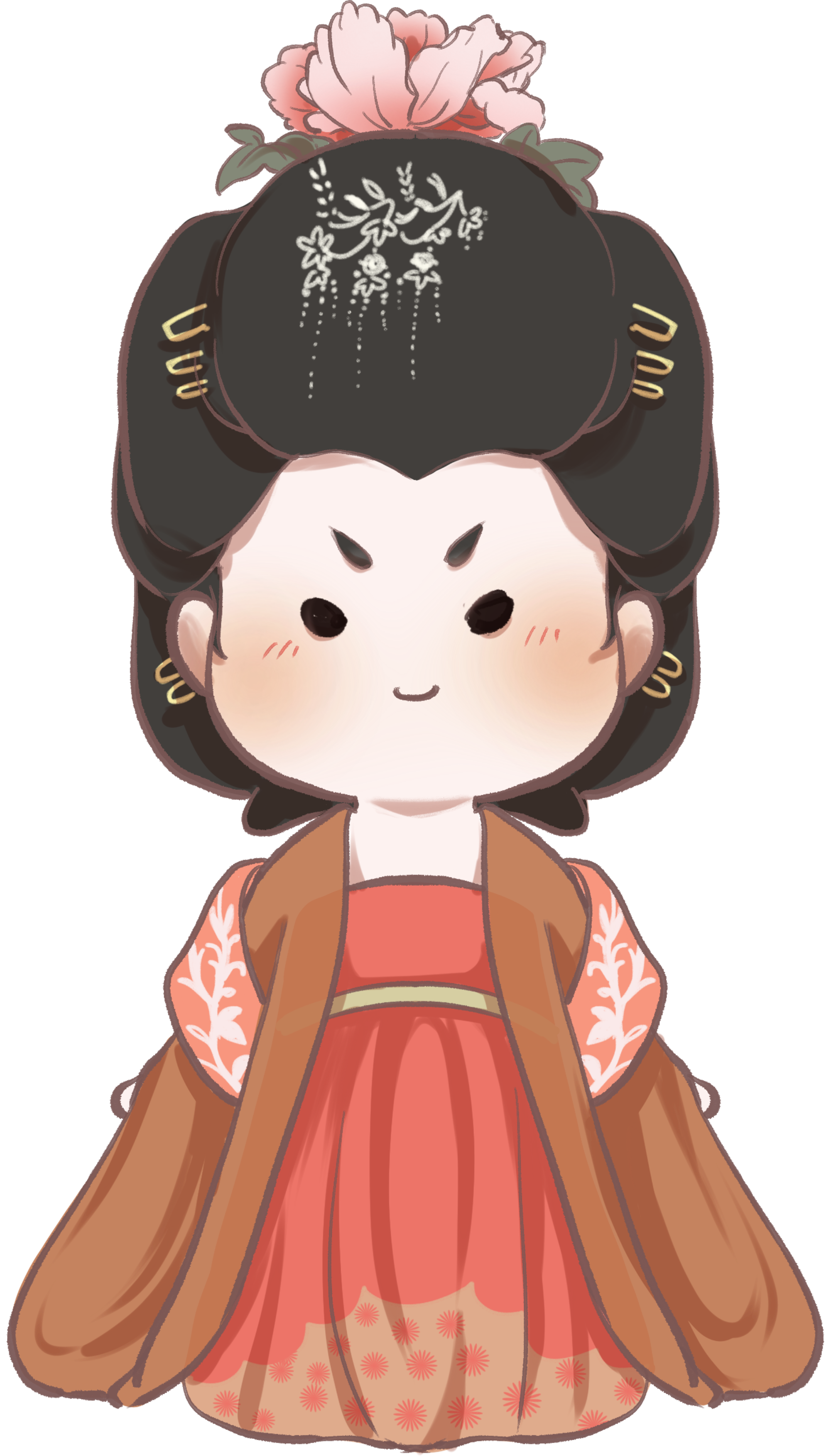} \textit{Hanfu-Bench}: A Multimodal Benchmark on Cross-Temporal Cultural Understanding and Transcreation}
\newcommand{\eqcontrib}{$^\dagger$}
\author{
 \textbf{Li Zhou\textsuperscript{1}\eqcontrib\thanks{Corresponding author}},
 \textbf{Lutong Yu\textsuperscript{1}\eqcontrib},
 \textbf{Dongchu Xie\textsuperscript{1}},
 \textbf{Shaohuan Cheng\textsuperscript{3}},
 \textbf{Wenyan Li\textsuperscript{4}},
 \textbf{Haizhou Li\textsuperscript{1,2}}\\
 \textsuperscript{1}The Chinese University of Hong Kong, Shenzhen  \textsuperscript{2}Shenzhen Research Institute of Big Data,\\
 \textsuperscript{3}Chengdu Technological University
 \textsuperscript{4}University of Copenhagen
\\
 \small{
   \href{lizhou21@cuhk.edu.cn, lutongyu@link.cuhk.edu.cn}{lizhou21@cuhk.edu.cn, lutongyu@link.cuhk.edu.cn}
 }
}

\begin{document}
\maketitle
\def\thefootnote{\eqcontrib}\footnotetext{Equal contribution.}\def\thefootnote{\arabic{footnote}}
\begin{abstract}
Culture is a rich and dynamic domain that evolves across both geography and time. However, existing studies on cultural understanding with vision-language models (VLMs) primarily emphasize geographic diversity, often overlooking the critical temporal dimensions.
To bridge this gap, we introduce Hanfu-Bench, a novel, expert-curated multimodal dataset. Hanfu, a traditional garment spanning ancient Chinese dynasties, serves as a representative cultural heritage that reflects the profound temporal aspects of Chinese culture while remaining highly popular in Chinese contemporary society.
Hanfu-Bench comprises two core tasks: 
cultural visual understanding and cultural image transcreation.
The former task examines temporal-cultural feature recognition based on single- or multi-image inputs through multiple-choice visual question answering, while the latter focuses on transforming traditional attire into modern designs through cultural element inheritance and modern context adaptation.
Our evaluation shows that closed VLMs perform comparably to non-experts on visual cutural understanding but fall short by 10\% to human experts, while open VLMs lags further behind non-experts. For the transcreation task, multi-faceted human evaluation indicates that the best-performing model achieves a success rate of only 42\%. Our benchmark provides an essential testbed, revealing significant challenges in this new direction of temporal cultural understanding and creative adaptation.
\footnote{Following \citet{jacovi-etal-2023-stop}, the Hanfu-Bench dataset is publicly available at \href{https://huggingface.co/datasets/lizhou21/Hanfu-Bench}{lizhou21/Hanfu-Bench} under the \text{CC BY-NC-SA 4.0 License}. The code details are freely available for reuse at \href{https://github.com/hlt-cuhksz/TemporalCulture}{hlt-cuhksz/TemporalCulture}.}

\end{abstract}
\section{Introduction}
\begin{figure}[t]
    \centering
    \includegraphics[width=0.85\linewidth]{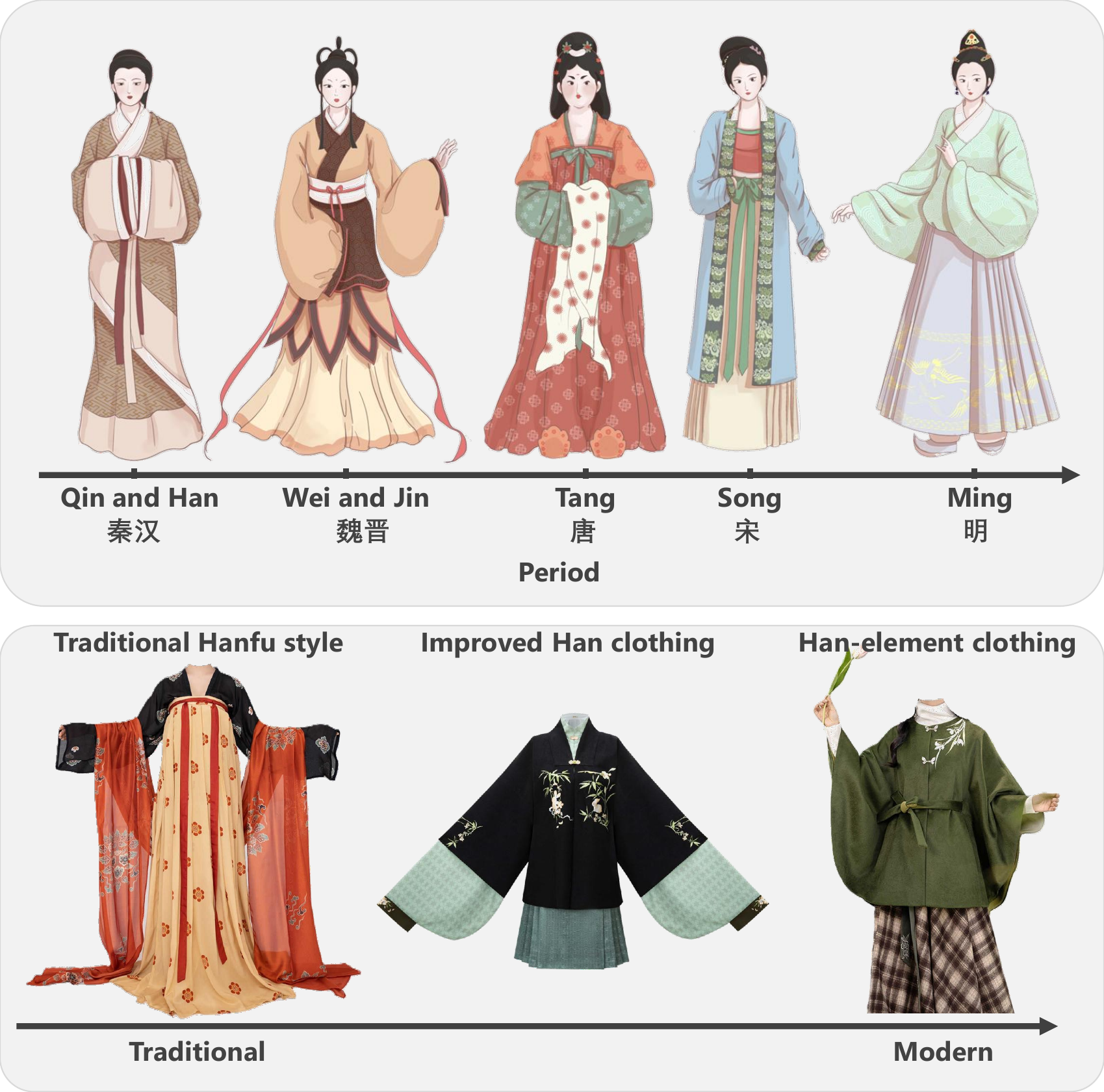}
    \caption{\textit{Top}: Hanfu styles from five distinct periods of Chinese history, illustrating variations in structure and design and tracing the evolution of traditional Chinese attire over time. \textit{Bottom}: The transformation of traditional Hanfu into modern clothing through cultural adaptation.
}
    \label{fig:intro}
\end{figure}
Traditional Chinese Hanfu exhibits distinct
characteristics in collar styles, sleeve designs, garment structures, and layering combinations across different historical periods~\cite{yu2024han, liu2022historical, Cai202404Research, KangZhaoNing2024}, as illustrated in Figure~\ref{fig:intro}.\footnote{The top part was hand-drawn by one of our authors, who is familiar with Hanfu. } 
For example, Hanfu of the Qin-Han period predominantly adhered to the \textit{Shenyi} (\begin{CJK}{UTF8}{gbsn}\xpinyin*{深衣}\end{CJK}) one-piece robe system, distinguished by its \textit{Raojin} (\begin{CJK}{UTF8}{gbsn}\xpinyin*{绕襟}\end{CJK}) construction—a wrapped-collar design with diagonally overlapping garment panels forming a cross-collar closure ~\cite{KangZhaoNing2024}.
Tang dynasty fashion featured the \textit{Pibo} (\begin{CJK}{UTF8}{gbsn}\xpinyin*{披帛}\end{CJK})—a long, lightweight silk scarf draped asymmetrically over robes—serves as a quintessential outer accessory~\cite{Liu2024digital}, while Ming dynasty clothing prominently showcased the \textit{Mamian Qun} (\begin{CJK}{UTF8}{gbsn}\xpinyin*{马面裙}\end{CJK}) as the characteristic skirt style~\cite{bao2025analysis, lin2023Study}.
These temporal variations of Hanfu highlight the intrinsic complexity of cultural knowledge, requiring an understanding spanning both historical and aesthetic dimensions.
This poses significant challenges for visual-language models (VLMs)~\cite{chang2024survey, li2025survey, li2025benchmark} in capturing culture-specific features across different historical periods.



However, existing research on cultural visual understanding primarily emphasizes geographic diversity~\cite{yin-etal-2021-broaden, 10.1145/3581783.3611994, NEURIPS2022_5474d9d4, nayak-etal-2024-benchmarking, schneider-sitaram-2024-m5, romero2025cvqa}, using broad regional~\cite{yin-etal-2021-broaden}, national~\cite{nayak-etal-2024-benchmarking, romero2025cvqa}, or sub-national~\cite{koto-etal-2024-indoculture, li-etal-2024-foodieqa} levels as cultural proxies, where temporal dynamics and historical evolution are underexplored~\cite{durham2020cultural}.

To bridge this gap, we introduce \textbf{Hanfu-Bench}, a manually curated multimodal dataset of traditional Chinese Hanfu spanning multiple dynasties, offering a cross-temporal perspective to evaluate models' ability to understand and apply temporal-cultural features. This dataset is specifically designed with two challenging tasks: \textbf{cultural visual understanding} and \textbf{cultural image transcreation}.
The first task assesses VLMs' ability to comprehend the temporal-cultural features of Hanfu through multiple-choice visual question answering (VQA). The second task evaluates their capacity to generate novel clothing designs by integrating ancient aesthetics into modern contexts via image translation.
The overall dataset construction and evaluation framework of Hanfu-Bench is shown in Figure~\ref{fig:pipeline}.


\textbf{Cultural visual understanding} involves question answering based on single- or multi-image inputs. Benchmarking with five advanced VLMs reveals that closed VLMs match non-expert human performance but fall short of experts, while open VLMs underperform even non-experts. Notably, while better at single-image tasks, VLMs struggle the most with multi-image VQA where humans excel. 


\textbf{Cultural image transcreation} in our work is a cross-temporal cultural adaptation task that examines whether the generative VLMs are capable of transforming traditional Hanfu images into modern designs while preserving their cultural essence. We evaluate on this task using a cascaded framework and measure the transcreation successful rate through human evaluations across six proposed dimensions.
The best-performing model achieves a success rate of only 42\%.



\textbf{Hanfu-Bench} underscores the limitations of current VLMs in capturing cultural nuances and temporal dynamics. By offering a comprehensive testbed, it provides valuable insights for future advancements. Models equipped with these capabilities can contribute to cultural heritage preservation~\cite{jin2022fluid, zou2023research, zhang2024from, bu2025investigation}, historical education~\cite{park2025ask, zhu2025exploring}, and innovative creative applications such as transcreation, which bridges the past and present through innovation.

\section{Related work}

As (visual)-language models become globally accessible, concerns about their biases in cultural values and knowledge have grown~\cite{cao-etal-2023-assessing, pawar2024survey, xu-etal-2025-self, zhou-etal-2025-mapo, bui-etal-2025-multi3hate}.
In the multimodal domain, research focuses on exploring the capabilities of models, 
from understanding cultural diversity to applying this knowledge through cultural adaptation.



\subsection{Cultural Visual Understanding}

To evaluate how VLMs interpret culturally diverse content, various datasets and benchmarks focus on tasks like visual question answering~\cite{nayak-etal-2024-benchmarking, romero2025cvqa, xu2025tcc}, image captioning~\cite{kadaoui2025jeem}, and reasoning~\cite{liu-etal-2021-visually}, revealing significant performance gaps across cultural and linguistic settings~\cite{khanuja-etal-2024-image, khanuja2024towards, mukherjee2025crossroads}. Geographic diversity datasets, such as GDVCR~\cite{yin-etal-2021-broaden}, CulturalVQA~\cite{nayak-etal-2024-benchmarking}, CVQA~\cite{romero2025cvqa}, CultureVerse~\cite{liu2025culturevlm}, and Worldcuisines~\cite{winata-etal-2025-worldcuisines}, show that VLMs struggle with culture-specific tasks and geo-diverse reasoning, especially in non-Western regions. Additionally, VLMs fail to capture fine-grained cultural nuances across regions, such as Arabic-speaking countries~\cite{kadaoui2025jeem}, Southeast Asia~\cite{urailertprasert-etal-2024-sea}, or Chinese regions~\cite{li-etal-2024-foodieqa}. Our study is the \texttt{first} to investigate cross-temporal cultural knowledge understanding, focusing on how VLMs interpret cultural evolution over time.

\begin{figure*}[t]
    \centering
    \includegraphics[width=0.95\linewidth]{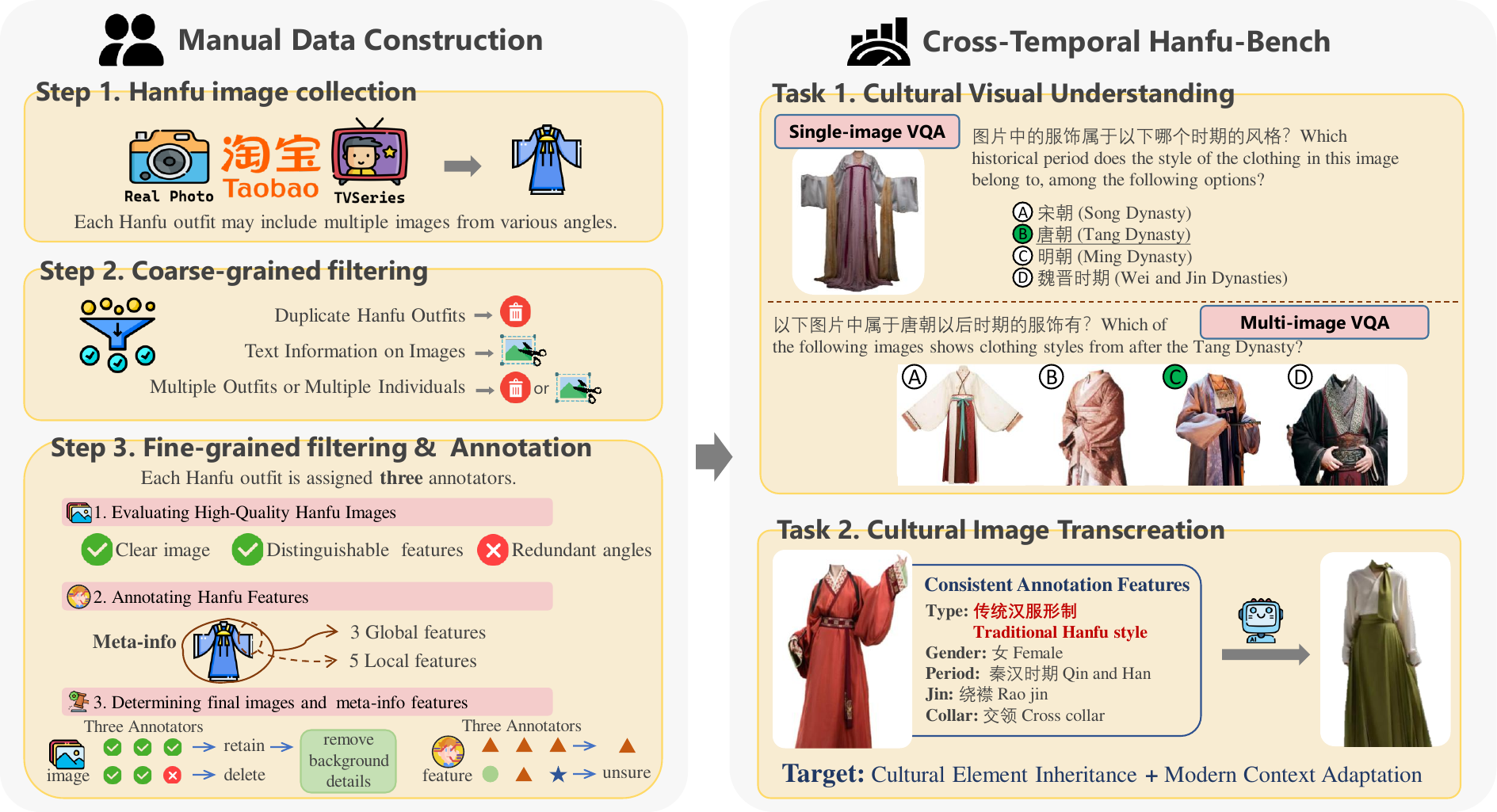}
    \caption{Data construction pipeline for Hanfu-Bench and task illustration. Left: our data construction pipeline includes image collection, filtering, and human annotation. Right: Hanfu-Bench includes tasks of both 1) cultural visual understanding and 2) cultural image transcreation.}
    \label{fig:pipeline}
\end{figure*}

\subsection{Cultural Adaptation}
Cultural adaptation, a culture-related downstream task, is often viewed as a macro-level translation process that adjusts concepts to fit the values and contexts of different cultures~\cite{liu2024culturally}. Recent studies address adaptation across multiple modalities, including text-to-text, text-to-image, and image-to-image translation. In text-based adaptation, tasks include recipe translation between Chinese and English-speaking cuisines~\cite{cao2024cultural} and synthesizing datasets for low-resource cultures~\cite{putri-etal-2024-llm}. For text-to-image tasks, CultDiff~\cite{bayramli2025diffusion} evaluates diffusion models' ability to generate culturally specific images across ten countries, while MosAIG~\cite{bhalerao2025multi} focuses on generating multicultural images that represent diverse cultures. In image-to-image translation, ~\citet{shin2024generative} use diffusion models to adapt dish compositions to different culinary styles, and ~\cite{khanuja-etal-2024-image} propose image transcreation to enhance cultural relevance. Our work extends image-to-image translation to cross-temporal cultural image transcreation, transforming traditional cultural elements into modern contexts.

\section{Data Collection and Annotation}
Hanfu, with its rich history, reflects diverse styles and details that evolved across eras. Our dataset collection includes three phases: (1) collecting Hanfu images of diverse styles from various periods; (2) image filtering to ensure quality and diversity; (3) expert annotation of temporal-cultural features with multiple granularity.



\subsection{Image collection}

We collect Hanfu images from both online and offline sources. 
For the online collection, it includes two categories: screenshots from Chinese television series and product images sourced from online Hanfu retailers.\footnote{We gathered the product images from online retailers at Taobao: \url{https://www.taobao.com/}.} 
Specifically, we use a web scraping framework to collect the product images, employing Selenium\footnote{\url{https://www.selenium.dev/}} for browser automation to simulate user interactions on Taobao. Other Hanfu image screenshots are collected manually.
The offline collection includes unique Hanfu images, captured in real-life settings, contributed by members of a Hanfu Club at a university.
To cover a wide range of diverse Hanfu styles spanning multiple dynasties, our selection includes 7 popular Hanfu retailers (chosen based on social media recommendations or high sales), alongside 7 critically acclaimed television series set in various Chinese dynasties. Specific sources are detailed in Appendix~\ref{app:data_source}.
We initially collected 902 Hanfu outfits, some with multiple images captured from various angles. 



\subsection{Image Filtering}
To curate a dataset of Hanfu images suitable for effectively evaluating VLMs, we implement a two-stage filtering process: coarse-grained and fine-grained filtering.
Coarse-grained filtering includes 1) deduplication on Hanfu outfits; 2) removing text elements to prevent text inference; 3) removing or cropping background subjects when multiple individuals are present in the image. 
Fine-grained filtering takes place during Hanfu feature annotation. Annotators are instructed to first evaluate images against three specific criteria before proceeding with detailed feature annotation: sufficient clarity, distinct and prominent features, and unique perspectives relative to other images of the same outfit. At this step, each hanfu outfit is reviewed by three different annotators, and only the images approved by all three annotators are included in the final dataset. To prevent non-Hanfu background information from interfering with model evaluations, all other irrelevant elements are removed manually.

\subsection{Expert Annotation}



Based on existing literature and through consultations with Hanfu experts, we identify eight key visual features for human annotation. These include three global features of the overall outfit, and five local features specifying clothing details.
 
The global features are: \textit{Type} (categorized as Traditional Hanfu, Improved Han clothing, or Han-element clothing),\footnote{Although sourced from high-quality Hanfu retailers, some designs include modern adaptations, aligning with the objectives of Task 2 in our work. Only outfits within these three categories are retained.} \textit{Period} (identifying the dynasty for Traditional Hanfu), and \textit{Gender} (indicating the intended or suitable gender). 
The local features focus on specific components of the outfit, including \textit{Sleeve}, \textit{Collar}, \textit{Jin} (\begin{CJK}{UTF8}{gbsn}\xpinyin*{襟}\end{CJK}, similar to lapels), \textit{Bottoms}, and \textit{Outerwear}. 
For each local feature, annotators are asked to choose from predefined options. They are encouraged to provide text descriptions for unique features not covered by these options and may also select ``none'' or ``unsure'' where appropriate.
We recruit nine annotators with expertise in Hanfu, comprising four authors of this study and five members of a Hanfu club. To ensure thorough and consistent assessments, each outfit is evaluated by three annotators. Detailed guidelines (Appendix~\ref{app:annotation_instruction}) are provided to enhance annotation reliability, including comprehensive explanations of each feature. The meta-information for a Hanfu outfit consists of feature values where all three annotators reached uniform agreement. We mark the feature value as ``unsure" when agreement could not be reached. 


In total, 496 sets of Hanfu are retained, comprising 1,192 images, with an average of 4.74 identified features per Hanfu.
The annotation platform and detailed distribution of annotated features, are provided in the Appendix~\ref{app:annotation_interface} and~\ref{app:dataset_statistics}.










\section{Task 1: Cultural Visual Understanding}
\label{sec:vqa}

\begin{table*}[t]
\centering
\scalebox{0.82}{
\begin{tabular}{@{}lrrrrrrrrr@{}}
\toprule
\multicolumn{1}{l|}{}                                & \textbf{Type} & \textbf{Gender} & \textbf{Period} & \textbf{Sleeve} & \textbf{Jin} & \textbf{Collar} & \textbf{Bottoms} & \multicolumn{1}{r|}{\textbf{Outerwear}} & \textbf{Overall} \\ \midrule
\multicolumn{10}{c}{\textbf{SVQA}}                                                                                                                                                                                                     \\ \midrule
\multicolumn{1}{l|}{\textbf{Count}}                  & 217           & 485             & 138             & 121             & 291          & 183             & 169              & \multicolumn{1}{r|}{117}                & 1721             \\ \midrule

\multicolumn{1}{l|}{\textbf{MiniCPM-V-2.6}}          & 70.51         & 90.31           & 44.93           & 51.24           & 65.98        & 49.73           & 68.05            & \multicolumn{1}{r|}{49.57}              & 68.04            \\
\multicolumn{1}{l|}{\textbf{Qwen2.5-VL-7B-Instruct}} & 74.65         & 88.25           & 42.75           & 63.64           & 40.89        & 53.55           & 41.42            & \multicolumn{1}{r|}{36.75}              & 61.36            \\
\multicolumn{1}{l|}{\textbf{InternVL2.5}}            & 78.34         & 94.23           & 44.93           & 66.12           & 78.01        & 55.74           & 43.79            & \multicolumn{1}{r|}{49.57}              & 71.47            \\
\multicolumn{1}{l|}{\textbf{GPT-4o}}                 & 74.65         & 96.49           & 73.19           & 70.25           & 61.17        & 59.02           & 88.76            & \multicolumn{1}{r|}{72.65}              & 77.69            \\
\multicolumn{1}{l|}{\textbf{Doubao-1.5-V}}           & 79.72         & 94.02           & 76.81           & 67.77           & 82.47        & 71.04           & 92.90             & \multicolumn{1}{r|}{61.54}              & 82.28            \\ \midrule
\multicolumn{10}{c}{\textbf{MVQA}}                                                                                                                                                                                                      \\ \midrule
\multicolumn{1}{l|}{\textbf{Count}}                  & 288           & 642             & 374             & 159             & 385          & 240             & 224              & \multicolumn{1}{r|}{153}                & 2465             \\ \midrule
\multicolumn{1}{l|}{\textbf{MiniCPM-V-2.6}}          & 21.53	     &59.03         	&22.99	         &25.79         	&18.70	      &26.67	        &34.38	          &\multicolumn{1}{r|}{33.99}	              &33.79            \\
\multicolumn{1}{l|}{\textbf{Qwen2.5-VL-7B-Instruct}} & 38.19         & 89.25           & 46.26           & 34.59           & 32.21        & 52.08           & 54.02            & \multicolumn{1}{r|}{52.29}              & 55.21            \\
\multicolumn{1}{l|}{\textbf{InternVL2.5}}            & 32.64         & 73.05           & 34.49           & 19.50            & 29.35        & 33.33           & 38.39            & \multicolumn{1}{r|}{37.91}              & 43.00            \\
\multicolumn{1}{l|}{\textbf{GPT-4o}}                 & 54.86         & 94.08           & 63.10            & 59.75           & 47.53        & 67.50            & 75.89            & \multicolumn{1}{r|}{69.28}              & 69.53            \\
\multicolumn{1}{l|}{\textbf{Doubao-1.5-V}}           & 46.72         & 94.95           & 83.71           & 100.00             & 62.40         & 70.46           & 74.88            & \multicolumn{1}{r|}{71.72}              & 75.95            \\ \bottomrule
\end{tabular}}
\caption{Comparison of SVQA and MVQA performance: multi-image VQA and understanding of temporal-cultural features (excluding Gender) pose greater challenges, especially for the open-weights VLMs.}
\label{tab:vqa_result}
\end{table*}

\subsection{VQA Tasks and Formualtion}

We introduce two types of multiple-choice visual question-answering tasks for temporal-cultural understanding: single-image visual question answering (SVQA) and multiple-image visual question answering (MVQA).
SVQA centers on features within a single Hanfu image, aiming to evaluate the model’s ability to recognize and interpret cultural elements from visual cues. In comparison, the MVQA task presents multiple Hanfu images simultaneously, requiring the model to identify and differentiate distinctive features across images.



\paragraph{Visual Question Formulation}
We developed a rule-based pipeline to automatically generate questions for both SVQA and MVQA tasks. For all questions, answer candidates are selected to ensure (1) exactly one correct answer exists and (2) no duplicate options appear. For SVQA, we create question templates across eight feature categories from our annotations, with answer candidates drawn from the annotated features. For MVQA, we designed templates to generate questions about features across image groups. For example, as shown in Figure~\ref{fig:pipeline}, when asking "Which image belongs to a period after Tang Dynasty?", we filter images based on their ``period" annotations to meet our answer selection criteria. 
The complete set of base questions and the corresponding attribute values are provided in Appendix~\ref{app:question templates}.

\subsection{Experimental Setup}
\label{subsec:setup}
We evaluate the SVQA and MVQA tasks using five state-of-the-art vision-language models capable of handling multi-image inputs. These include three open-source models—MiniCPM-V 2.6~\cite{yao2024minicpm}, Qwen2.5-VL-7B-Instruct~\cite{qwen2.5-VL}, and InternVL2.5~\cite{chen2024internvl}—and two closed-source models, GPT-4o~\cite{openai2024gpt4technicalreport} and Doubao1.5-vision.\footnote{\url{https://www.volcengine.com/}}
Based on the base questions from our VQA framework, we design five task-specific Chinese instructions that define task roles, describe the tasks, and outline output requirements. All prompt settings are provided in Appendix~\ref{app:vqa_prompt}.


\begin{figure}
    \centering
    \includegraphics[width=1\linewidth]{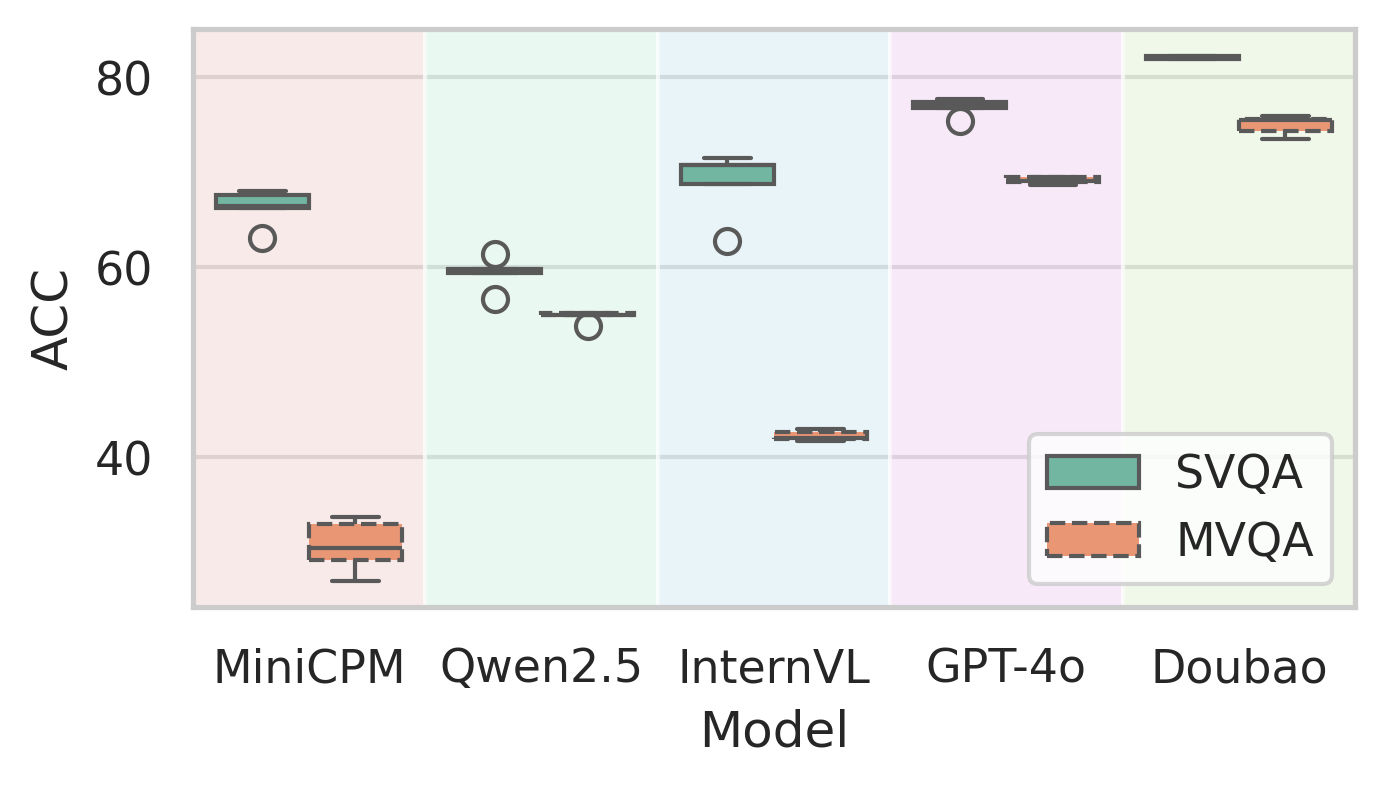}
    \caption{Accuracy of SVQA and MVQA tasks across four different prompts: SVQA is more sensitive to prompt variations.}
    \label{fig:prompt_sensitive}
\end{figure}




\subsection{Results and Analysis}



\paragraph{Multi-Image VQA is More Difficult for VLMs.}
Table~\ref{tab:vqa_result} presents the accuracy of the best-performing prompt across various VLMs for SVQA and MVQA tasks. Overall, both closed-source and open-source models perform significantly worse on MVQA than SVQA, highlighting challenges in integrating information across multiple images and understanding the temporal and cultural characteristics of Hanfu.
Among all models, the closed-source Doubao-1.5-V achieves the best overall performance. For open-source models, InternVL2.5 performs best on SVQA, while Qwen2.5-VL-7B-Instruct leads on MVQA.
VLMs perform better on non-cultural attributes, such as Gender, but struggle with recognizing culturally significant features that vary over time and space. Figure~\ref{fig:prompt_sensitive} illustrates the sensitivity of the models to different prompts across the two tasks, showing that MVQA is less affected by prompt variations than SVQA, likely due to its stronger reliance on visual content over textual instructions.
However, specific role definitions, such as in Prompt 3 (``You are an expert with deep knowledge of traditional clothing culture and its modern adaptations''), can introduce bias, lowering accuracy for attributes like \textit{Type} and causing overall performance drops, as reflected in the SVQA outlier in Figure~\ref{fig:prompt_sensitive}.

\begin{table}[]
\centering
\scalebox{0.9}{
\begin{tabular}{@{}l|rr@{}}
\toprule
                                & \textbf{SVQA} & \textbf{MVQA} \\ \midrule
\textbf{Non-Expert Human}     & 57.92         &  64.58        \\
\textbf{Expert Human}         & 77.50         &  83.54       \\ \midrule
\textbf{MiniCPM-V-2.6}          & 52.50         &  27.50            \\
\textbf{Qwen2.5-VL-7B-Instruct} & 48.33         & 48.12          \\
\textbf{InternVL2.5}            & 50.62        & 36.46          \\
\textbf{GPT-4o}                 & 66.25        &  66.67           \\
\textbf{Doubao-1.5-V}           & 70.00         &  74.48            \\ \bottomrule
\end{tabular}}
\caption{Human vs. VLM performance on SVQA and MVQA with balanced question subsets: humans perform better on MVQA.}
\label{tab:human}
\end{table}

\begin{figure}[t]
    \centering
    \includegraphics[width=0.9\linewidth]{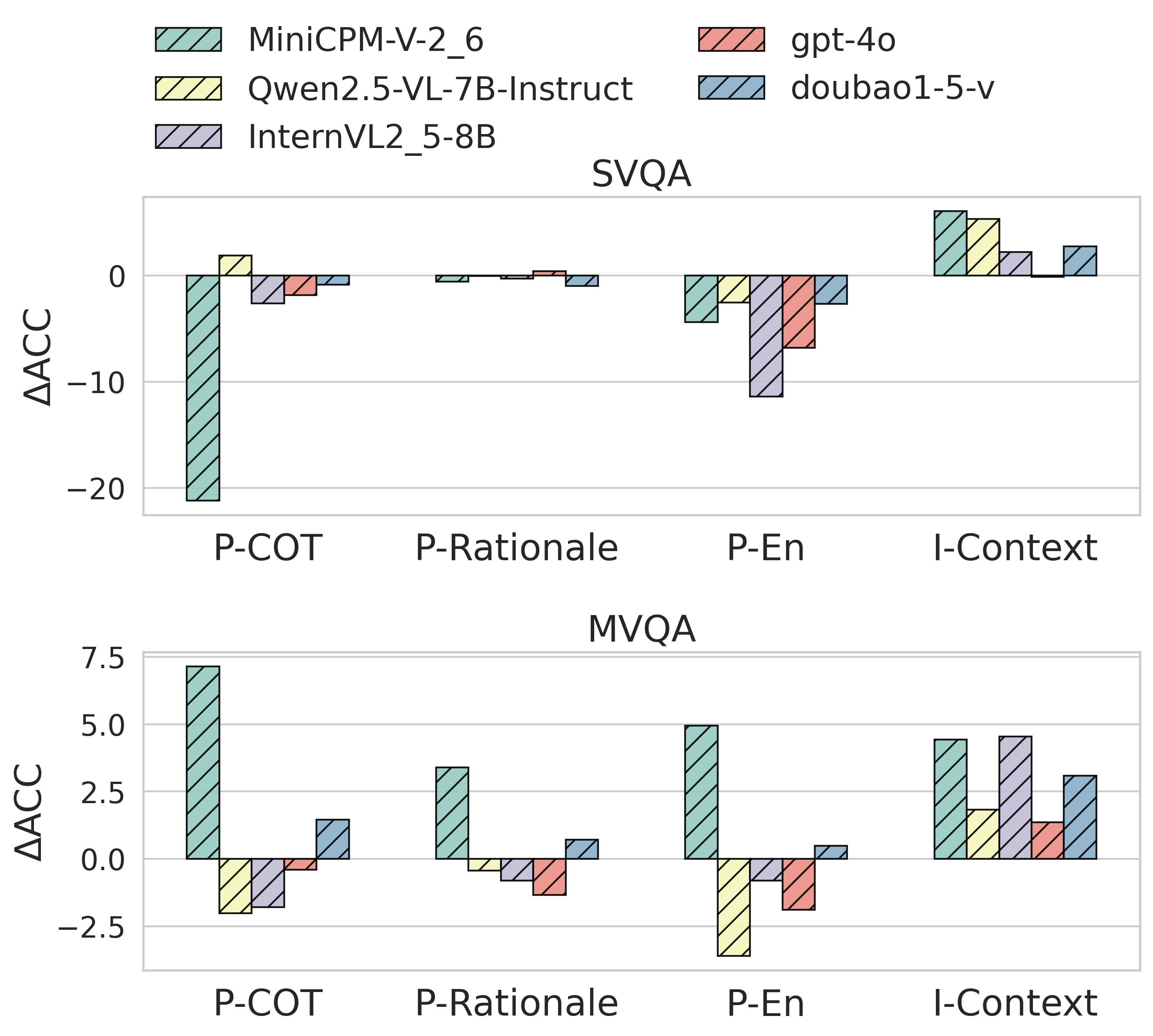}
    \caption{Performance changes of VLMs with different prompt types and image contexts compared to the base Chinese prompt. \textit{P-COT}: step-by-step reasoning instructions; \textit{P-Rationale}: requiring answer rationales; \textit{P-En}: English prompts; \textit{I-context}: images with background retained.}
    \label{fig:variable}
\end{figure}


\paragraph{Humans Perform Better on Multi-Image VQA.}
To compare the performance of VLMs with humans, we select 20 questions per feature type, ensuring a balanced distribution of true labels. This results in two subsets of 160 questions each for the SVQA and MVQA tasks. Six evaluators—three familiar with Hanfu and three unfamiliar—participate in the evaluation, with results presented in Table~\ref{tab:human}.
Notably, humans perform better on MVQA tasks than on SVQA tasks, regardless of whether they are experts or non-experts—contrasting sharply with VLMs’ performance. Interviews with the evaluators reveal that when presented with multiple images, they reason by identifying feature differences across images. This reasoning proves particularly effective when the four images are not entirely dissimilar, enabling meaningful comparisons. Additionally, evaluators acquire knowledge about Hanfu styles through the answering process and use previously answered questions to infer responses to new ones. 
These findings expose current VLMs’ limitations in integrating cross-image information and contextual learning, underscoring a substantial gap between human reasoning and model capabilities in MVQA.
Moreover, while open-source VLMs outperform non-expert humans, they do not surpass experts; closed-source VLMs perform worse than non-experts, further illustrating the challenges VLMs face in achieving human-level understanding and reasoning in this domain.

\paragraph{Image Background is Helpful for Reasoning.}
To investigate the impact of prompt phrasing, language use, and image backgrounds, we design three prompt variations and one image variation: P-COT adds step-by-step reasoning instructions, P-Rationale requires a rationale for answers, P-En uses English prompts, and I-context retains background information in input images. 
The accuracy differences compared to the base Chinese Prompt 1 are shown in Figure~\ref{fig:variable}.
We find that when images include additional non-clothing background information, all models show consistent performance improvements across tasks, particularly for global feature types and gender. However, for other feature predictions, the inclusion of background information introduces noise in some models, which disrupts accurate judgment.
Besides, when the prompt instructions are in English, all models show a decrease in consistency on SVQA. This suggests that models have a better understanding of traditional Chinese clothing descriptions in Chinese, likely due to challenges in aligning culturally specific terms between Chinese and English. In contrast, on MVQA, both MiniCPM-V 2.6 and Doubao-1.5-V show slight improvements with English prompts. This may be because in MVQA, only the question description includes specific cultural terms, minimizing the impact of language differences.
Additionally, while incorporating reasoning requirements in the prompt or asking VLMs to provide rationales for their answers is intended to enhance performance, its effect on VLMs remains unclear and does not result in significant improvements.









\begin{figure*}[ht]
    \centering
    \includegraphics[width=1\linewidth]{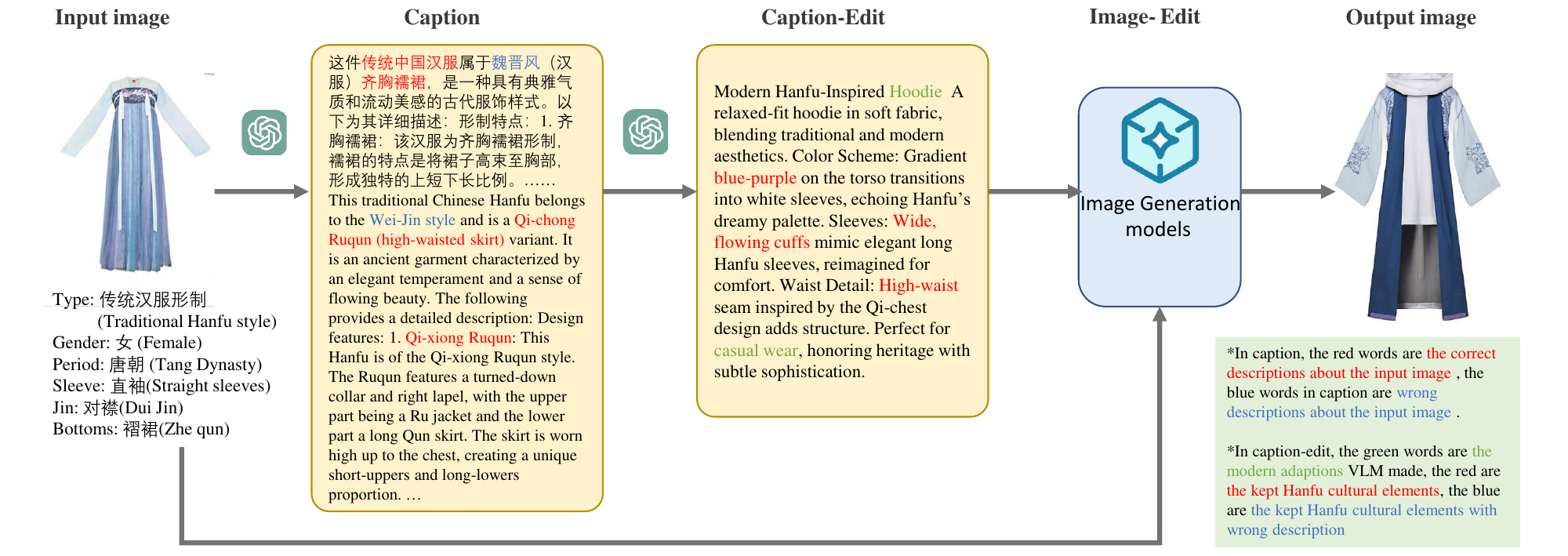}
    \caption{Cultural image transcreation pipeline and case study. VLMs face challenges in extracting and utilizing cultural elements. Modifications to extracted cultural elements can lead to loss or misjudgment of their original characteristics. In this case, wrongly altering the feature of straight sleeves (which is concise and straight) to wide and flowing cuffs causes the sleeve features in the generated image to deviate from the original hanfu image.}
    \label{fig:image-generation-pipeline}
\end{figure*}

\begin{table*}[]
\centering
\resizebox{\textwidth}{!}{
\begin{tabular}{@{}lll@{}}
\toprule
\textbf{ID} & \textbf{Question}                                                                                                                                    & \textbf{Property}    \\ \midrule
\textbf{C0} & Is there any visual change in the generated image compared to the original image? & Visual-change        \\
\textbf{C1} & Is the content of this image clothing?                                        & Semantic-equivalence \\
\textbf{C2} & Does the clothing in this image reflect naturally occurring in the life?                     & Naturalness          \\
\textbf{C3} &  Is the clothing in this image suitable for modern daily wear?            & Modern- adaptability \\
\textbf{C4} &  Does the generated image retain the Hanfu cultural elements from the original image?& Cultural-Inheritance \\
\textbf{C5} &  Does the clothing in this image match your aesthetic preferences?                & Attractiveness       \\ \bottomrule
\end{tabular}}
\caption{Questions asked for evaluating cultural image transcreation task.}
\label{tab:human_detail}
\end{table*}

\section{Task 2: Cultural Image Transcreation}

\subsection{Task Definition and Evaluation Metric}

We propose a cross-temporal cultural adaptation task aimed at transforming traditional Hanfu images into modern, wearable designs while preserving their cultural essence. Different from the image transcreation concept in \citet{khanuja-etal-2024-image}, which focuses on cross-regional(spatial) cultural adaptation for improved cultural relevance, our task emphasizes \textit{cross-temporal} 
image translation to achieve modern adaptation.
As shown in Figure~\ref{fig:intro}, the modern adaptive design of Hanfu closely aligns with the improved Han clothing and Han-element styles featured in the Hanfu-Bench collection, demonstrating the practical application of this adaptation process.
We employ a questionnaire-based design for human evaluation. In addition to retaining the three commonly used evaluation dimensions for image generation editing—\textbf{visual-change}, \textbf{semantic-equivalence}, and \textbf{naturalness}—we introduce three additional dimensions specifically tailored to the unique characteristics of our task: \textbf{modern-adaptability}, \textbf{cultural-inheritance}, and \textbf{attractiveness}.
Human evaluators are instructed to compare the original image with the generated image considering each of the six evaluation dimensions and rate on a 5-point scale, as detailed in Table~\ref{tab:human_detail}. A comprehensive explanation of these quantitative metrics is provided in Appendix~\ref{app:Quantitative Metrics of Task2}.


\subsection{Experimental Setup}


We implement a cascaded framework for this task, shown in Figure~\ref{fig:image-generation-pipeline}. We experiment with 50 representative traditional Hanfu images from Hanfu-Bench as inputs, which are carefully selected for their strong visual features. The framework first generates descriptive image captions for a given image. The caption is then refined through an editing stage to ensure the descriptions adapt to modern requirements. Finally, the edited caption guides image editing, resulting in a generated image of creative modern design that fuses Hanfu cultural elements.

For image captioning and caption editing, we utilize GPT-4o, known for its strong capabilities in cultural and visual understanding, as well as its proficiency in English.\footnote{During the caption editing stage, Chinese captions are translated into English to enhance compatibility with existing image generation models.} 
For image generation, we employ three models: Instruct-Pix2Pix~\cite{brooks2023instructpix2pixlearningfollowimage}, Stable-Diffusion v2-1-base (SD)~\cite{Rombach_2022_CVPR}, and SD-XL 1.0-base (SDXL)~\cite{podell2023sdxlimprovinglatentdiffusion}.
We recruited five experts from a Hanfu club as human evaluators to assess the quality of the 150 generated images from the three models. Evaluators were presented with two images side by side: the original image on the left and the corresponding generated image on the right. To ensure fair evaluation, the images are randomly presented without information about the generation model disclosed.

\begin{figure*}[ht]
    \centering
    \includegraphics[width=1\linewidth]{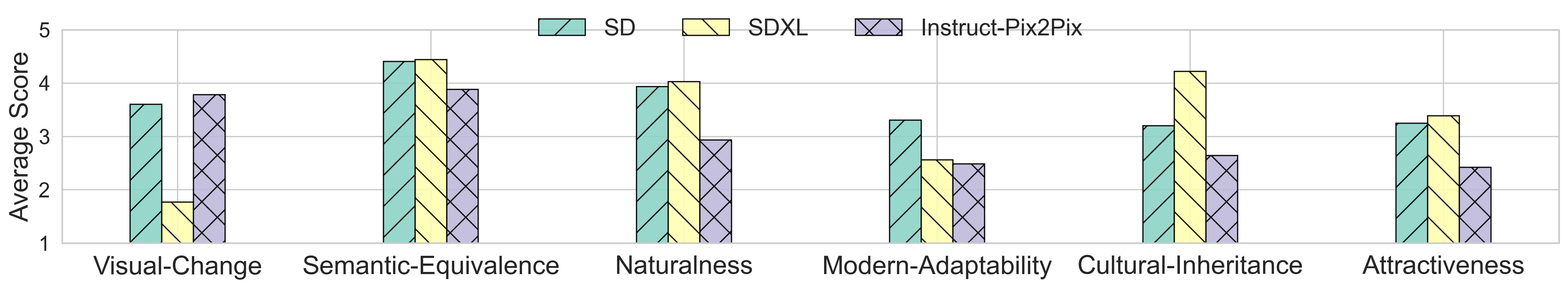}
    \caption{Comparison of the average scores across six quantitative metrics for the images of modern adaptive hanfu designs generated by three different models }
    \label{fig:model_performance_comparison}
\end{figure*}

\subsection{Results and Analysis}

\paragraph{Model Comparison of Quantitative Metrics}  
Figure~\ref{fig:model_performance_comparison} presents the average scores of six evaluation metrics for the three image generation models in generating modern adaptive Hanfu designs.
All models score low in \textbf{modern adaptability}, with only the generated images from the SD model exceed 3, indicating moderately suitable for modern daily wear. This suggests big challenges in modernizing Hanfu design: the VLMs may not correctly understand Hanfu features, making modernization difficult; or image generation models may struggle to apply cultural elements in modern contexts.
For \textbf{cultural inheritance}, SDXL scores relatively high (4.2) due to insufficient visual changes. The other two models achieve mid-range scores (around 3.0). This suggests that this framework has limited capability in identifying and preserving Hanfu cultural elements.
In \textbf{visual change}, SD and Instruct-Pix2Pix both exceed 3, while SDXL scores lower. SDXL tops in semantic equivalence (4.4) and naturalness (3.9), and attractiveness (3.3). This could be relevant to SDXL's focus on high-precision images, leading to high similarity in overall contours but some minor detail changes.
For \textbf{semantic equivalence}, all three models score around 4, showing no significant semantic ambiguity, with Instruct-Pix2Pix slightly lower than the other two models. And all models show average performance in \textbf{naturalness}, with SD and SDXL scoring around 4, while Instruct-Pix2Pix lags slightly behind at approximately 3. Its emphasis on local modifications based on text instructions may hinder its ability in generating realistic images compared to the other two models. This might also explain its scores below 3 in other dimensions.
In Figure~\ref{fig:result-task2} we present more qualitative results.


\paragraph{Tradition-to-modern transformation is challenging}

As our task is due to transforming traditional Hanfu images into modern, we prioritize three key evaluation metrics: semantic consistency of the generated images, which ensures the outputs serve as valid fashion reference; the modern adaptability of the design, indicating that the images should represent a contemporary garment; and cultural preservation, as our goal is to produce clothing with Hanfu cultural elements and characteristics. Consequently, images are regarded as compliant if they score at least 3 in each of the three dimensions: semantic equivalence (C1), modern adaptability (C3), and cultural inheritance (C4). Among the three models, SD achieves the highest success rate at 42.0\%, while Instruct-Pix2Pix performs the worst with only 8.0\%. Nevertheless, none of the models surpass the 50\% threshold for successful image generation, revealing the significant challenges in enhancing models' comprehension and utilization of cultural elements across different dimensions.

\section{Conclusion and Discussion}

In this paper, we present Hanfu-Bench, a manually curated multimodal dataset composed of traditional Chinese Hanfu designs spanning various dynasties.
This dataset serves as an essential resource for evaluating cultural understanding and creative adaptability of VLMs.
Through the two core tasks—cultural visual understanding and cultural image transcreation—our experiments reveal significant limitations of current models in capturing and distringuishing subtle temporal-cultural features and effectively adapt ancient designs into modern contexts.
This work highlights AI's critical role in promoting cultural heritage, bridging the understanding of tradition with innovative digital reinterpretation.

However, the effective application of AI in this domain also presents challenges. Models may inherit biases from imbalanced or modernized image sources, potentially distorting Hanfu’s traditional aesthetics and symbolic meanings~\cite{ma2023study}. This raises challenges of cultural fidelity in creative reinterpretation~\cite{tiribelli2024ethics}, especially when AI is applied to generate modern clothing designs inspired by Hanfu elements. Therefore, using diverse, culturally grounded datasets and ensuring expert involvement in validation are crucial. Furthermore, even advanced models like GPT-4o, despite their strong multimodal capabilities, often struggle with classical Chinese content and may produce hallucinated interpretations \cite{alawida2023comprehensive}. Future work should emphasize fairness-aware development, domain-specific fine-tuning, and close collaboration with cultural scholars to ensure AI supports preservation rather than misrepresentation of culture heritage.

\section*{Limitations}
While the proposed dataset is rich in cultural attributes, its reliance on expert annotations and the requirement for agreement among three annotators in determining feature values constrain its size and pose significant challenges to scalability and extensibility. This limitation underscores the trade-off between ensuring high-quality, culturally grounded data and the practical challenges of expanding the dataset to support broader applications.

Moreover, the performance of VLMs in image understanding may be influenced by objective factors such as the angle, clarity, and composition of the clothing images (e.g., full-body vs. half-body shots). However, these factors were not annotated during the initial dataset curation, which limits the study's ability to conduct an in-depth analysis of their impact on model performance. This omission highlights the importance of incorporating such metadata in future dataset iterations to enable more comprehensive evaluations.

Additionally, as parts of the dataset were sourced from the internet, there is a possibility that some models may have already encountered similar images during pretraining. This raises concerns about data leakage and the potential overestimation of model capabilities. 



\section*{Ethical Considerations}
This study explores the application of vision-language models in recognizing Hanfu from diverse image sources, revealing both potential and limitations. The dataset used includes publicly available images from e-commerce platforms and media sources, with acknowledgment of potential copyright implications; all images are used and shared strictly for academic research, with no commercial intent. 

Furthermore, our use of text-to-image models to generate new Hanfu images adapted for modern contexts introduces a potential risk: inaccurate cultural representation. Similar to all other image-generation tasks, these models can carry inherent biases towards certain features or cultures, which could unintentionally affect the authentic dissemination of Hanfu culture.

\section*{Dataset License Statement}
This dataset is licensed under the Creative Commons Attribution-NonCommercial-ShareAlike 4.0 International (CC BY-NC-SA 4.0) license, which requires users to provide appropriate attribution to the original author(s), restricts usage to non-commercial purposes, and mandates that any derivative works be distributed under the same license. Additionally, the dataset is strictly limited to academic research purposes, including activities related to education, academic research, and scholarly communication, such as writing academic papers, evaluating models, and conducting experiments. It may only be used for evaluation purposes and not for training models or systems. Non-academic non-commercial activities, such as personal projects or non-academic nonprofit initiatives, are not permitted. All users must ensure compliance with relevant institutional and national laws and regulations during usage.

\section*{Acknowledgments}
We are deeply grateful to all the volunteers for their generous contributions and efforts in providing high-quality annotations and evaluations. 
Specifically, we would like to thank the Han Society club at The Chinese University of Hong Kong, Shenzhen, for their valuable support.
We further acknowledge the contributions of Siyuan Wang, Lujia Zhang, Xinran Xu, Mingyuan Li, Xuanyu Wang, and Hanke Qiao for their meticulous work on Hanfu annotation. 
We also extend our gratitude to Anxiao He, Ziyan Liu, Yirui Guo, Li Chen, Mo Li, and Ruohao Han for participating in the expert and non-expert human evaluations for the Cultural Visual Understanding task. 
Finally, we appreciate the efforts of Chengyi Chen, Hanke Qiao, Chen Yang, Siyu Liu, and Yirui Guo in evaluating the generated images for the Cultural Image Transcreation task.

This work is supported by Shenzhen Science and Technology Program (Shenzhen Key Laboratory, Grant No. ZDSYS20230626091302006), Program for Guangdong Introducing Innovative and Entrepreneurial Teams (Grant No. 2023ZT10X044), Shenzhen Stability Science Program 2023, Shenzhen Key Lab of Multi-Modal Cognitive Computing, Shenzhen Science and Technology Research Fund (Fundamental Research Key Project, Grant No.JCYJ20220818103001002), the International Science and Technology Cooperation Center, Ministry of Science and Technology of China (under grant 2024YFE0203000), the Internal Project of Shenzhen Research Institute of Big Data.
Wenyan Li is supported by the Lundbeck Foundation (BrainDrugs grant: R279-2018-1145). 

\bibliography{custom,anthology}

\begin{thebibliography}{56}
\providecommand{\natexlab}[1]{#1}

\bibitem[{Alawida et~al.(2023)Alawida, Mejri, Mehmood, Chikhaoui, and Isaac~Abiodun}]{alawida2023comprehensive}
Moatsum Alawida, Sami Mejri, Abid Mehmood, Belkacem Chikhaoui, and Oludare Isaac~Abiodun. 2023.
\newblock A comprehensive study of chatgpt: advancements, limitations, and ethical considerations in natural language processing and cybersecurity.
\newblock \emph{Information}, 14(8):462.

\bibitem[{Bao and Guo(2025)}]{bao2025analysis}
Shuyi Bao and Tianjiao Guo. 2025.
\newblock An analysis of the traditional horse-face skirt structure and its modern design.
\newblock \emph{Journal of Silk}, 62(02):10--20.

\bibitem[{Bayramli et~al.(2025)Bayramli, Suleymanzade, An, Ahmad, Kim, Park, Thorne, and Oh}]{bayramli2025diffusion}
Zahra Bayramli, Ayhan Suleymanzade, Na~Min An, Huzama Ahmad, Eunsu Kim, Junyeong Park, James Thorne, and Alice Oh. 2025.
\newblock Diffusion models through a global lens: Are they culturally inclusive?
\newblock \emph{arXiv preprint arXiv:2502.08914}.

\bibitem[{Bhalerao et~al.(2025)Bhalerao, Yalamarty, Trinh, and Ignat}]{bhalerao2025multi}
Parth Bhalerao, Mounika Yalamarty, Brian Trinh, and Oana Ignat. 2025.
\newblock Multi-agent multimodal models for multicultural text to image generation.
\newblock \emph{arXiv preprint arXiv:2502.15972}.

\bibitem[{Brooks et~al.(2023)Brooks, Holynski, and Efros}]{brooks2023instructpix2pixlearningfollowimage}
Tim Brooks, Aleksander Holynski, and Alexei~A. Efros. 2023.
\newblock \href {https://arxiv.org/abs/2211.09800} {Instructpix2pix: Learning to follow image editing instructions}.
\newblock \emph{Preprint}, arXiv:2211.09800.

\bibitem[{Bu et~al.(2025)Bu, Wang, Wang, and Liu}]{bu2025investigation}
Fan Bu, Zheng Wang, Siyi Wang, and Ziyao Liu. 2025.
\newblock An investigation into value misalignment in llm-generated texts for cultural heritage.
\newblock \emph{arXiv preprint arXiv:2501.02039}.

\bibitem[{Bui et~al.(2025)Bui, Wense, and Lauscher}]{bui-etal-2025-multi3hate}
Minh~Duc Bui, Katharina Von~Der Wense, and Anne Lauscher. 2025.
\newblock \href {https://aclanthology.org/2025.naacl-long.490/} {{M}ulti$^3${H}ate: Multimodal, multilingual, and multicultural hate speech detection with vision{--}language models}.
\newblock In \emph{Proceedings of the 2025 Conference of the Nations of the Americas Chapter of the Association for Computational Linguistics: Human Language Technologies (Volume 1: Long Papers)}, pages 9714--9731, Albuquerque, New Mexico. Association for Computational Linguistics.

\bibitem[{Cai and Liu(2024)}]{Cai202404Research}
Yiming Cai and Anding Liu. 2024.
\newblock \href {https://doi.org/10.62051/ijsspa.v2n3.48} {Research on the evolution of ancient women’s wedding attire}.
\newblock \emph{International Journal of Social Sciences and Public Administration}, 2(3):342–350.

\bibitem[{Cao et~al.(2024)Cao, Kementchedjhieva, Cui, Karamolegkou, Zhou, Dare, Donatelli, and Hershcovich}]{cao2024cultural}
Yong Cao, Yova Kementchedjhieva, Ruixiang Cui, Antonia Karamolegkou, Li~Zhou, Megan Dare, Lucia Donatelli, and Daniel Hershcovich. 2024.
\newblock Cultural adaptation of recipes.
\newblock \emph{Transactions of the Association for Computational Linguistics}, 12:80--99.

\bibitem[{Cao et~al.(2023)Cao, Zhou, Lee, Cabello, Chen, and Hershcovich}]{cao-etal-2023-assessing}
Yong Cao, Li~Zhou, Seolhwa Lee, Laura Cabello, Min Chen, and Daniel Hershcovich. 2023.
\newblock \href {https://doi.org/10.18653/v1/2023.c3nlp-1.7} {Assessing cross-cultural alignment between {C}hat{GPT} and human societies: An empirical study}.
\newblock In \emph{Proceedings of the First Workshop on Cross-Cultural Considerations in NLP (C3NLP)}, pages 53--67, Dubrovnik, Croatia. Association for Computational Linguistics.

\bibitem[{Chang et~al.(2024)Chang, Wang, Wang, Wu, Yang, Zhu, Chen, Yi, Wang, Wang et~al.}]{chang2024survey}
Yupeng Chang, Xu~Wang, Jindong Wang, Yuan Wu, Linyi Yang, Kaijie Zhu, Hao Chen, Xiaoyuan Yi, Cunxiang Wang, Yidong Wang, et~al. 2024.
\newblock A survey on evaluation of large language models.
\newblock \emph{ACM transactions on intelligent systems and technology}, 15(3):1--45.

\bibitem[{Chen et~al.(2024)Chen, Wu, Wang, Su, Chen, Xing, Zhong, Zhang, Zhu, Lu et~al.}]{chen2024internvl}
Zhe Chen, Jiannan Wu, Wenhai Wang, Weijie Su, Guo Chen, Sen Xing, Muyan Zhong, Qinglong Zhang, Xizhou Zhu, Lewei Lu, et~al. 2024.
\newblock Internvl: Scaling up vision foundation models and aligning for generic visual-linguistic tasks.
\newblock In \emph{Proceedings of the IEEE/CVF Conference on Computer Vision and Pattern Recognition}, pages 24185--24198.

\bibitem[{Durham(2020)}]{durham2020cultural}
William~H Durham. 2020.
\newblock Cultural variation in time and space: the case for a populational theory of culture.
\newblock In \emph{Anthropology beyond culture}, pages 193--206. Routledge.

\bibitem[{Gaviria~Rojas et~al.(2022)Gaviria~Rojas, Diamos, Kini, Kanter, Janapa~Reddi, and Coleman}]{NEURIPS2022_5474d9d4}
William Gaviria~Rojas, Sudnya Diamos, Keertan Kini, David Kanter, Vijay Janapa~Reddi, and Cody Coleman. 2022.
\newblock \href {https://proceedings.neurips.cc/paper_files/paper/2022/file/5474d9d43c0519aa176276ff2c1ca528-Paper-Datasets_and_Benchmarks.pdf} {The dollar street dataset: Images representing the geographic and socioeconomic diversity of the world}.
\newblock In \emph{Advances in Neural Information Processing Systems}, volume~35, pages 12979--12990. Curran Associates, Inc.

\bibitem[{Jacovi et~al.(2023)Jacovi, Caciularu, Goldman, and Goldberg}]{jacovi-etal-2023-stop}
Alon Jacovi, Avi Caciularu, Omer Goldman, and Yoav Goldberg. 2023.
\newblock \href {https://doi.org/10.18653/v1/2023.emnlp-main.308} {Stop uploading test data in plain text: Practical strategies for mitigating data contamination by evaluation benchmarks}.
\newblock In \emph{Proceedings of the 2023 Conference on Empirical Methods in Natural Language Processing}, pages 5075--5084, Singapore. Association for Computational Linguistics.

\bibitem[{Jin and Liu(2022)}]{jin2022fluid}
Pei Jin and Yi~Liu. 2022.
\newblock \href {https://doi.org/10.1016/j.teler.2022.100022} {Fluid space: Digitisation of cultural heritage and its media dissemination}.
\newblock \emph{Telematics and Informatics Reports}, 8:100022.

\bibitem[{Kadaoui et~al.(2025)Kadaoui, Atwany, Al-Ali, Mohamed, Mekky, Tilga, Fedorova, Artemova, Aldarmaki, and Kementchedjhieva}]{kadaoui2025jeem}
Karima Kadaoui, Hanin Atwany, Hamdan Al-Ali, Abdelrahman Mohamed, Ali Mekky, Sergei Tilga, Natalia Fedorova, Ekaterina Artemova, Hanan Aldarmaki, and Yova Kementchedjhieva. 2025.
\newblock Jeem: Vision-language understanding in four arabic dialects.
\newblock \emph{arXiv preprint arXiv:2503.21910}.

\bibitem[{Kang et~al.(2024)Kang, Zhao, and Ning}]{KangZhaoNing2024}
Kai Kang, Kaiwei Zhao, and Feiyan Ning. 2024.
\newblock \href {https://doi.org/10.54691/dk2tt897} {Research on hanfu flower decoration under the background of cultural confidence}.
\newblock \emph{Scientific Journal Of Humanities and Social Sciences}, 6(8):66–70.

\bibitem[{Khanuja et~al.(2024{\natexlab{a}})Khanuja, Iyer, He, and Neubig}]{khanuja2024towards}
Simran Khanuja, Vivek Iyer, Claire He, and Graham Neubig. 2024{\natexlab{a}}.
\newblock Towards automatic evaluation for image transcreation.
\newblock \emph{arXiv preprint arXiv:2412.13717}.

\bibitem[{Khanuja et~al.(2024{\natexlab{b}})Khanuja, Ramamoorthy, Song, and Neubig}]{khanuja-etal-2024-image}
Simran Khanuja, Sathyanarayanan Ramamoorthy, Yueqi Song, and Graham Neubig. 2024{\natexlab{b}}.
\newblock \href {https://doi.org/10.18653/v1/2024.emnlp-main.573} {An image speaks a thousand words, but can everyone listen? on image transcreation for cultural relevance}.
\newblock In \emph{Proceedings of the 2024 Conference on Empirical Methods in Natural Language Processing}, pages 10258--10279, Miami, Florida, USA. Association for Computational Linguistics.

\bibitem[{Koto et~al.(2024)Koto, Mahendra, Aisyah, and Baldwin}]{koto-etal-2024-indoculture}
Fajri Koto, Rahmad Mahendra, Nurul Aisyah, and Timothy Baldwin. 2024.
\newblock \href {https://doi.org/10.1162/tacl_a_00726} {{I}ndo{C}ulture: Exploring geographically influenced cultural commonsense reasoning across eleven {I}ndonesian provinces}.
\newblock \emph{Transactions of the Association for Computational Linguistics}, 12:1703--1719.

\bibitem[{Li et~al.(2024)Li, Zhang, Li, Peng, Tang, Zhou, Zhang, Hu, Yuan, S{\o}gaard, Hershcovich, and Elliott}]{li-etal-2024-foodieqa}
Wenyan Li, Crystina Zhang, Jiaang Li, Qiwei Peng, Raphael Tang, Li~Zhou, Weijia Zhang, Guimin Hu, Yifei Yuan, Anders S{\o}gaard, Daniel Hershcovich, and Desmond Elliott. 2024.
\newblock \href {https://doi.org/10.18653/v1/2024.emnlp-main.1063} {{F}oodie{QA}: A multimodal dataset for fine-grained understanding of {C}hinese food culture}.
\newblock In \emph{Proceedings of the 2024 Conference on Empirical Methods in Natural Language Processing}, pages 19077--19095, Miami, Florida, USA. Association for Computational Linguistics.

\bibitem[{Li et~al.(2025{\natexlab{a}})Li, Wu, Du, Liu, Nghiem, and Shi}]{li2025survey}
Zongxia Li, Xiyang Wu, Hongyang Du, Fuxiao Liu, Huy Nghiem, and Guangyao Shi. 2025{\natexlab{a}}.
\newblock A survey of state of the art large vision language models: Alignment, benchmark, evaluations and challenges.

\bibitem[{Li et~al.(2025{\natexlab{b}})Li, Wu, Du, Nghiem, and Shi}]{li2025benchmark}
Zongxia Li, Xiyang Wu, Hongyang Du, Huy Nghiem, and Guangyao Shi. 2025{\natexlab{b}}.
\newblock \href {https://arxiv.org/abs/2501.02189} {Benchmark evaluations, applications, and challenges of large vision language models: A survey}.
\newblock \emph{Preprint}, arXiv:2501.02189.

\bibitem[{Lin and Niu(2023)}]{lin2023Study}
Pengfei Lin and Li~Niu. 2023.
\newblock Study on horse face skirts unearthed from tombs of the wang luo family in changzhou in the ming dynasty.
\newblock \emph{Journal of Silk}, 60(07):135--142.

\bibitem[{Liu et~al.(2024{\natexlab{a}})Liu, Gurevych, and Korhonen}]{liu2024culturally}
Chen~Cecilia Liu, Iryna Gurevych, and Anna Korhonen. 2024{\natexlab{a}}.
\newblock Culturally aware and adapted nlp: A taxonomy and a survey of the state of the art.
\newblock \emph{arXiv preprint arXiv:2406.03930}.

\bibitem[{Liu et~al.(2024{\natexlab{b}})Liu, Cui, and Wang}]{Liu2024digital}
Chunxiao Liu, RongRong Cui, and Zhicheng Wang. 2024{\natexlab{b}}.
\newblock \href {https://doi.org/10.3390/jtaer19020069} {Digital virtual simulation for cultural clothing restoration: Case study of tang dynasty mural ‘diplomatic envoys’ from crown prince zhang huai’s tomb}.
\newblock \emph{Journal of Theoretical and Applied Electronic Commerce Research}, 19(2):1358--1391.

\bibitem[{Liu et~al.(2021)Liu, Bugliarello, Ponti, Reddy, Collier, and Elliott}]{liu-etal-2021-visually}
Fangyu Liu, Emanuele Bugliarello, Edoardo~Maria Ponti, Siva Reddy, Nigel Collier, and Desmond Elliott. 2021.
\newblock \href {https://doi.org/10.18653/v1/2021.emnlp-main.818} {Visually grounded reasoning across languages and cultures}.
\newblock In \emph{Proceedings of the 2021 Conference on Empirical Methods in Natural Language Processing}, pages 10467--10485, Online and Punta Cana, Dominican Republic. Association for Computational Linguistics.

\bibitem[{Liu et~al.(2022)Liu, Zhou, and Zhu}]{liu2022historical}
Kaixuan Liu, Shunmuzi Zhou, and Chun Zhu. 2022.
\newblock Historical changes of chinese costumes from the perspective of archaeology.
\newblock \emph{Heritage Science}, 10(1):205.

\bibitem[{Liu et~al.(2025)Liu, Jin, Li, Wong, Wen, Sun, Chen, Xie, and Wang}]{liu2025culturevlm}
Shudong Liu, Yiqiao Jin, Cheng Li, Derek~F Wong, Qingsong Wen, Lichao Sun, Haipeng Chen, Xing Xie, and Jindong Wang. 2025.
\newblock Culturevlm: Characterizing and improving cultural understanding of vision-language models for over 100 countries.
\newblock \emph{arXiv preprint arXiv:2501.01282}.

\bibitem[{Ma(2023)}]{ma2023study}
Yongfeng Ma. 2023.
\newblock A study of ethical issues in natural language processing with artificial intelligence.
\newblock \emph{Journal of Computer Science and Technology Studies}, 5(1):52--56.

\bibitem[{Ma et~al.(2023)Ma, Pan, Wu, Cheng, Zhang, Huang, and Chen}]{10.1145/3581783.3611994}
Zheng Ma, Mianzhi Pan, Wenhan Wu, Kanzhi Cheng, Jianbing Zhang, Shujian Huang, and Jiajun Chen. 2023.
\newblock \href {https://doi.org/10.1145/3581783.3611994} {Food-500 cap: A fine-grained food caption benchmark for evaluating vision-language models}.
\newblock In \emph{Proceedings of the 31st ACM International Conference on Multimedia}, MM '23, page 5674–5685, New York, NY, USA. Association for Computing Machinery.

\bibitem[{Mukherjee et~al.(2025)Mukherjee, Zhu, and Anastasopoulos}]{mukherjee2025crossroads}
Anjishnu Mukherjee, Ziwei Zhu, and Antonios Anastasopoulos. 2025.
\newblock Crossroads of continents: Automated artifact extraction for cultural adaptation with large multimodal models.
\newblock In \emph{2025 IEEE/CVF Winter Conference on Applications of Computer Vision (WACV)}, pages 1755--1764. IEEE.

\bibitem[{Nayak et~al.(2024)Nayak, Jain, Awal, Reddy, Steenkiste, Hendricks, Stanczak, and Agrawal}]{nayak-etal-2024-benchmarking}
Shravan Nayak, Kanishk Jain, Rabiul Awal, Siva Reddy, Sjoerd~Van Steenkiste, Lisa~Anne Hendricks, Karolina Stanczak, and Aishwarya Agrawal. 2024.
\newblock \href {https://doi.org/10.18653/v1/2024.emnlp-main.329} {Benchmarking vision language models for cultural understanding}.
\newblock In \emph{Proceedings of the 2024 Conference on Empirical Methods in Natural Language Processing}, pages 5769--5790, Miami, Florida, USA. Association for Computational Linguistics.

\bibitem[{OpenAI et~al.(2024)OpenAI, Achiam, Adler, Agarwal, Ahmad, Akkaya, Aleman, Almeida, Altenschmidt, Altman, Anadkat, Avila, Babuschkin, Balaji, Balcom, Baltescu, Bao, Bavarian, Belgum, Bello, Berdine, Bernadett-Shapiro, Berner, Bogdonoff, Boiko, Boyd, Brakman, Brockman, Brooks, Brundage, Button, Cai, Campbell, Cann, Carey, Carlson, Carmichael, Chan, Chang, Chantzis, Chen, Chen, Chen, Chen, Chen, Chess, Cho, Chu, Chung, Cummings, Currier, Dai, Decareaux, Degry, Deutsch, Deville, Dhar, Dohan, Dowling, Dunning, Ecoffet, Eleti, Eloundou, Farhi, Fedus, Felix, Fishman, Forte, Fulford, Gao, Georges, Gibson, Goel, Gogineni, Goh, Gontijo-Lopes, Gordon, Grafstein, Gray, Greene, Gross, Gu, Guo, Hallacy, Han, Harris, He, Heaton, Heidecke, Hesse, Hickey, Hickey, Hoeschele, Houghton, Hsu, Hu, Hu, Huizinga, Jain, Jain, Jang, Jiang, Jiang, Jin, Jin, Jomoto, Jonn, Jun, Kaftan, Łukasz Kaiser, Kamali, Kanitscheider, Keskar, Khan, Kilpatrick, Kim, Kim, Kim, Kirchner, Kiros, Knight, Kokotajlo, Łukasz Kondraciuk,
  Kondrich, Konstantinidis, Kosic, Krueger, Kuo, Lampe, Lan, Lee, Leike, Leung, Levy, Li, Lim, Lin, Lin, Litwin, Lopez, Lowe, Lue, Makanju, Malfacini, Manning, Markov, Markovski, Martin, Mayer, Mayne, McGrew, McKinney, McLeavey, McMillan, McNeil, Medina, Mehta, Menick, Metz, Mishchenko, Mishkin, Monaco, Morikawa, Mossing, Mu, Murati, Murk, Mély, Nair, Nakano, Nayak, Neelakantan, Ngo, Noh, Ouyang, O'Keefe, Pachocki, Paino, Palermo, Pantuliano, Parascandolo, Parish, Parparita, Passos, Pavlov, Peng, Perelman, de~Avila Belbute~Peres, Petrov, de~Oliveira~Pinto, Michael, Pokorny, Pokrass, Pong, Powell, Power, Power, Proehl, Puri, Radford, Rae, Ramesh, Raymond, Real, Rimbach, Ross, Rotsted, Roussez, Ryder, Saltarelli, Sanders, Santurkar, Sastry, Schmidt, Schnurr, Schulman, Selsam, Sheppard, Sherbakov, Shieh, Shoker, Shyam, Sidor, Sigler, Simens, Sitkin, Slama, Sohl, Sokolowsky, Song, Staudacher, Such, Summers, Sutskever, Tang, Tezak, Thompson, Tillet, Tootoonchian, Tseng, Tuggle, Turley, Tworek, Uribe, Vallone,
  Vijayvergiya, Voss, Wainwright, Wang, Wang, Wang, Ward, Wei, Weinmann, Welihinda, Welinder, Weng, Weng, Wiethoff, Willner, Winter, Wolrich, Wong, Workman, Wu, Wu, Wu, Xiao, Xu, Yoo, Yu, Yuan, Zaremba, Zellers, Zhang, Zhang, Zhao, Zheng, Zhuang, Zhuk, and Zoph}]{openai2024gpt4technicalreport}
OpenAI, Josh Achiam, Steven Adler, Sandhini Agarwal, Lama Ahmad, Ilge Akkaya, Florencia~Leoni Aleman, Diogo Almeida, Janko Altenschmidt, Sam Altman, Shyamal Anadkat, Red Avila, Igor Babuschkin, Suchir Balaji, Valerie Balcom, Paul Baltescu, Haiming Bao, Mohammad Bavarian, Jeff Belgum, Irwan Bello, Jake Berdine, Gabriel Bernadett-Shapiro, Christopher Berner, Lenny Bogdonoff, Oleg Boiko, Madelaine Boyd, Anna-Luisa Brakman, Greg Brockman, Tim Brooks, Miles Brundage, Kevin Button, Trevor Cai, Rosie Campbell, Andrew Cann, Brittany Carey, Chelsea Carlson, Rory Carmichael, Brooke Chan, Che Chang, Fotis Chantzis, Derek Chen, Sully Chen, Ruby Chen, Jason Chen, Mark Chen, Ben Chess, Chester Cho, Casey Chu, Hyung~Won Chung, Dave Cummings, Jeremiah Currier, Yunxing Dai, Cory Decareaux, Thomas Degry, Noah Deutsch, Damien Deville, Arka Dhar, David Dohan, Steve Dowling, Sheila Dunning, Adrien Ecoffet, Atty Eleti, Tyna Eloundou, David Farhi, Liam Fedus, Niko Felix, Simón~Posada Fishman, Juston Forte, Isabella Fulford, Leo
  Gao, Elie Georges, Christian Gibson, Vik Goel, Tarun Gogineni, Gabriel Goh, Rapha Gontijo-Lopes, Jonathan Gordon, Morgan Grafstein, Scott Gray, Ryan Greene, Joshua Gross, Shixiang~Shane Gu, Yufei Guo, Chris Hallacy, Jesse Han, Jeff Harris, Yuchen He, Mike Heaton, Johannes Heidecke, Chris Hesse, Alan Hickey, Wade Hickey, Peter Hoeschele, Brandon Houghton, Kenny Hsu, Shengli Hu, Xin Hu, Joost Huizinga, Shantanu Jain, Shawn Jain, Joanne Jang, Angela Jiang, Roger Jiang, Haozhun Jin, Denny Jin, Shino Jomoto, Billie Jonn, Heewoo Jun, Tomer Kaftan, Łukasz Kaiser, Ali Kamali, Ingmar Kanitscheider, Nitish~Shirish Keskar, Tabarak Khan, Logan Kilpatrick, Jong~Wook Kim, Christina Kim, Yongjik Kim, Jan~Hendrik Kirchner, Jamie Kiros, Matt Knight, Daniel Kokotajlo, Łukasz Kondraciuk, Andrew Kondrich, Aris Konstantinidis, Kyle Kosic, Gretchen Krueger, Vishal Kuo, Michael Lampe, Ikai Lan, Teddy Lee, Jan Leike, Jade Leung, Daniel Levy, Chak~Ming Li, Rachel Lim, Molly Lin, Stephanie Lin, Mateusz Litwin, Theresa Lopez, Ryan
  Lowe, Patricia Lue, Anna Makanju, Kim Malfacini, Sam Manning, Todor Markov, Yaniv Markovski, Bianca Martin, Katie Mayer, Andrew Mayne, Bob McGrew, Scott~Mayer McKinney, Christine McLeavey, Paul McMillan, Jake McNeil, David Medina, Aalok Mehta, Jacob Menick, Luke Metz, Andrey Mishchenko, Pamela Mishkin, Vinnie Monaco, Evan Morikawa, Daniel Mossing, Tong Mu, Mira Murati, Oleg Murk, David Mély, Ashvin Nair, Reiichiro Nakano, Rajeev Nayak, Arvind Neelakantan, Richard Ngo, Hyeonwoo Noh, Long Ouyang, Cullen O'Keefe, Jakub Pachocki, Alex Paino, Joe Palermo, Ashley Pantuliano, Giambattista Parascandolo, Joel Parish, Emy Parparita, Alex Passos, Mikhail Pavlov, Andrew Peng, Adam Perelman, Filipe de~Avila Belbute~Peres, Michael Petrov, Henrique~Ponde de~Oliveira~Pinto, Michael, Pokorny, Michelle Pokrass, Vitchyr~H. Pong, Tolly Powell, Alethea Power, Boris Power, Elizabeth Proehl, Raul Puri, Alec Radford, Jack Rae, Aditya Ramesh, Cameron Raymond, Francis Real, Kendra Rimbach, Carl Ross, Bob Rotsted, Henri Roussez,
  Nick Ryder, Mario Saltarelli, Ted Sanders, Shibani Santurkar, Girish Sastry, Heather Schmidt, David Schnurr, John Schulman, Daniel Selsam, Kyla Sheppard, Toki Sherbakov, Jessica Shieh, Sarah Shoker, Pranav Shyam, Szymon Sidor, Eric Sigler, Maddie Simens, Jordan Sitkin, Katarina Slama, Ian Sohl, Benjamin Sokolowsky, Yang Song, Natalie Staudacher, Felipe~Petroski Such, Natalie Summers, Ilya Sutskever, Jie Tang, Nikolas Tezak, Madeleine~B. Thompson, Phil Tillet, Amin Tootoonchian, Elizabeth Tseng, Preston Tuggle, Nick Turley, Jerry Tworek, Juan Felipe~Cerón Uribe, Andrea Vallone, Arun Vijayvergiya, Chelsea Voss, Carroll Wainwright, Justin~Jay Wang, Alvin Wang, Ben Wang, Jonathan Ward, Jason Wei, CJ~Weinmann, Akila Welihinda, Peter Welinder, Jiayi Weng, Lilian Weng, Matt Wiethoff, Dave Willner, Clemens Winter, Samuel Wolrich, Hannah Wong, Lauren Workman, Sherwin Wu, Jeff Wu, Michael Wu, Kai Xiao, Tao Xu, Sarah Yoo, Kevin Yu, Qiming Yuan, Wojciech Zaremba, Rowan Zellers, Chong Zhang, Marvin Zhang, Shengjia
  Zhao, Tianhao Zheng, Juntang Zhuang, William Zhuk, and Barret Zoph. 2024.
\newblock \href {https://arxiv.org/abs/2303.08774} {Gpt-4 technical report}.
\newblock \emph{Preprint}, arXiv:2303.08774.

\bibitem[{Park et~al.(2025)Park, Song, Seo, Kim, and Suh}]{park2025ask}
Kieun Park, Hyungwoo Song, Seungbae Seo, Junghwan Kim, and Bongwon Suh. 2025.
\newblock " ask sir oliver ingham": Llm-based social simulations for history education.
\newblock In \emph{Proceedings of the Extended Abstracts of the CHI Conference on Human Factors in Computing Systems}, pages 1--13.

\bibitem[{Pawar et~al.(2024)Pawar, Park, Jin, Arora, Myung, Yadav, Haznitrama, Song, Oh, and Augenstein}]{pawar2024survey}
Siddhesh Pawar, Junyeong Park, Jiho Jin, Arnav Arora, Junho Myung, Srishti Yadav, Faiz~Ghifari Haznitrama, Inhwa Song, Alice Oh, and Isabelle Augenstein. 2024.
\newblock Survey of cultural awareness in language models: Text and beyond.
\newblock \emph{arXiv preprint arXiv:2411.00860}.

\bibitem[{Podell et~al.(2023)Podell, English, Lacey, Blattmann, Dockhorn, Müller, Penna, and Rombach}]{podell2023sdxlimprovinglatentdiffusion}
Dustin Podell, Zion English, Kyle Lacey, Andreas Blattmann, Tim Dockhorn, Jonas Müller, Joe Penna, and Robin Rombach. 2023.
\newblock \href {https://arxiv.org/abs/2307.01952} {Sdxl: Improving latent diffusion models for high-resolution image synthesis}.
\newblock \emph{Preprint}, arXiv:2307.01952.

\bibitem[{Putri et~al.(2024)Putri, Haznitrama, Adhista, and Oh}]{putri-etal-2024-llm}
Rifki~Afina Putri, Faiz~Ghifari Haznitrama, Dea Adhista, and Alice Oh. 2024.
\newblock \href {https://doi.org/10.18653/v1/2024.emnlp-main.1145} {Can {LLM} generate culturally relevant commonsense {QA} data? case study in {I}ndonesian and {S}undanese}.
\newblock In \emph{Proceedings of the 2024 Conference on Empirical Methods in Natural Language Processing}, pages 20571--20590, Miami, Florida, USA. Association for Computational Linguistics.

\bibitem[{Rombach et~al.(2022)Rombach, Blattmann, Lorenz, Esser, and Ommer}]{Rombach_2022_CVPR}
Robin Rombach, Andreas Blattmann, Dominik Lorenz, Patrick Esser, and Bj\"orn Ommer. 2022.
\newblock High-resolution image synthesis with latent diffusion models.
\newblock In \emph{Proceedings of the IEEE/CVF Conference on Computer Vision and Pattern Recognition (CVPR)}, pages 10684--10695.

\bibitem[{Romero et~al.(2025)Romero, Lyu, Wibowo, G{\'o}ngora, Mandal, Purkayastha, Ortiz-Barajas, Cueva, Baek, Jeong et~al.}]{romero2025cvqa}
David Romero, Chenyang Lyu, Haryo Wibowo, Santiago G{\'o}ngora, Aishik Mandal, Sukannya Purkayastha, Jesus-German Ortiz-Barajas, Emilio Cueva, Jinheon Baek, Soyeong Jeong, et~al. 2025.
\newblock Cvqa: Culturally-diverse multilingual visual question answering benchmark.
\newblock \emph{Advances in Neural Information Processing Systems}, 37:11479--11505.

\bibitem[{Schneider and Sitaram(2024)}]{schneider-sitaram-2024-m5}
Florian Schneider and Sunayana Sitaram. 2024.
\newblock \href {https://doi.org/10.18653/v1/2024.findings-emnlp.250} {M5 {--} a diverse benchmark to assess the performance of large multimodal models across multilingual and multicultural vision-language tasks}.
\newblock In \emph{Findings of the Association for Computational Linguistics: EMNLP 2024}, pages 4309--4345, Miami, Florida, USA. Association for Computational Linguistics.

\bibitem[{Shin et~al.(2024)Shin, Sridhar, Sampson, and Narayanan}]{shin2024generative}
Philip~Wootaek Shin, Ajay~Narayanan Sridhar, Jack Sampson, and Vijaykrishnan Narayanan. 2024.
\newblock A generative exploration of cuisine transfer.
\newblock In \emph{Proceedings of the IEEE/CVF Conference on Computer Vision and Pattern Recognition}, pages 3732--3740.

\bibitem[{Team(2025)}]{qwen2.5-VL}
Qwen Team. 2025.
\newblock \href {https://qwenlm.github.io/blog/qwen2.5-vl/} {Qwen2.5-vl}.

\bibitem[{Tiribelli et~al.(2024)Tiribelli, Pansoni, Frontoni, and Giovanola}]{tiribelli2024ethics}
Simona Tiribelli, Sofia Pansoni, Emanuele Frontoni, and Benedetta Giovanola. 2024.
\newblock Ethics of artificial intelligence for cultural heritage: Opportunities and challenges.
\newblock \emph{IEEE Transactions on Technology and Society}.

\bibitem[{Urailertprasert et~al.(2024)Urailertprasert, Limkonchotiwat, Suwajanakorn, and Nutanong}]{urailertprasert-etal-2024-sea}
Norawit Urailertprasert, Peerat Limkonchotiwat, Supasorn Suwajanakorn, and Sarana Nutanong. 2024.
\newblock \href {https://doi.org/10.18653/v1/2024.alvr-1.15} {{SEA}-{VQA}: {S}outheast {A}sian cultural context dataset for visual question answering}.
\newblock In \emph{Proceedings of the 3rd Workshop on Advances in Language and Vision Research (ALVR)}, pages 173--185, Bangkok, Thailand. Association for Computational Linguistics.

\bibitem[{Winata et~al.(2025)Winata, Hudi, Irawan, Anugraha, Putri, Yutong, Nohejl, Prathama, Ousidhoum, Amriani, Rzayev, Das, Pramodya, Adila, Wilie, Mawalim, Lam, Abolade, Chersoni, Santus, Ikhwantri, Kuwanto, Zhao, Wibowo, Lovenia, Cruz, Putra, Myung, Susanto, Machin, Zhukova, Anugraha, Adilazuarda, Santosa, Limkonchotiwat, Dabre, Audino, Cahyawijaya, Zhang, Salim, Zhou, Gui, Adelani, Lee, Okada, Purwarianti, Aji, Watanabe, Wijaya, Oh, and Ngo}]{winata-etal-2025-worldcuisines}
Genta~Indra Winata, Frederikus Hudi, Patrick~Amadeus Irawan, David Anugraha, Rifki~Afina Putri, Wang Yutong, Adam Nohejl, Ubaidillah~Ariq Prathama, Nedjma Ousidhoum, Afifa Amriani, Anar Rzayev, Anirban Das, Ashmari Pramodya, Aulia Adila, Bryan Wilie, Candy~Olivia Mawalim, Cheng~Ching Lam, Daud Abolade, Emmanuele Chersoni, Enrico Santus, Fariz Ikhwantri, Garry Kuwanto, Hanyang Zhao, Haryo~Akbarianto Wibowo, Holy Lovenia, Jan Christian~Blaise Cruz, Jan Wira~Gotama Putra, Junho Myung, Lucky Susanto, Maria Angelica~Riera Machin, Marina Zhukova, Michael Anugraha, Muhammad~Farid Adilazuarda, Natasha~Christabelle Santosa, Peerat Limkonchotiwat, Raj Dabre, Rio~Alexander Audino, Samuel Cahyawijaya, Shi-Xiong Zhang, Stephanie~Yulia Salim, Yi~Zhou, Yinxuan Gui, David~Ifeoluwa Adelani, En-Shiun~Annie Lee, Shogo Okada, Ayu Purwarianti, Alham~Fikri Aji, Taro Watanabe, Derry~Tanti Wijaya, Alice Oh, and Chong-Wah Ngo. 2025.
\newblock \href {https://aclanthology.org/2025.naacl-long.167/} {{W}orld{C}uisines: A massive-scale benchmark for multilingual and multicultural visual question answering on global cuisines}.
\newblock In \emph{Proceedings of the 2025 Conference of the Nations of the Americas Chapter of the Association for Computational Linguistics: Human Language Technologies (Volume 1: Long Papers)}, pages 3242--3264, Albuquerque, New Mexico. Association for Computational Linguistics.

\bibitem[{Xu et~al.(2025{\natexlab{a}})Xu, Wang, Zhang, Zhou, Li, Yuan, Li, Zhou, Wang, Zhang et~al.}]{xu2025tcc}
Pengju Xu, Yan Wang, Shuyuan Zhang, Xuan Zhou, Xin Li, Yue Yuan, Fengzhao Li, Shunyuan Zhou, Xingyu Wang, Yi~Zhang, et~al. 2025{\natexlab{a}}.
\newblock Tcc-bench: Benchmarking the traditional chinese culture understanding capabilities of mllms.
\newblock \emph{arXiv preprint arXiv:2505.11275}.

\bibitem[{Xu et~al.(2025{\natexlab{b}})Xu, Leng, Yu, and Xiong}]{xu-etal-2025-self}
Shaoyang Xu, Yongqi Leng, Linhao Yu, and Deyi Xiong. 2025{\natexlab{b}}.
\newblock \href {https://aclanthology.org/2025.naacl-long.350/} {Self-pluralising culture alignment for large language models}.
\newblock In \emph{Proceedings of the 2025 Conference of the Nations of the Americas Chapter of the Association for Computational Linguistics: Human Language Technologies (Volume 1: Long Papers)}, pages 6859--6877, Albuquerque, New Mexico. Association for Computational Linguistics.

\bibitem[{Yao et~al.(2024)Yao, Yu, Zhang, Wang, Cui, Zhu, Cai, Li, Zhao, He et~al.}]{yao2024minicpm}
Yuan Yao, Tianyu Yu, Ao~Zhang, Chongyi Wang, Junbo Cui, Hongji Zhu, Tianchi Cai, Haoyu Li, Weilin Zhao, Zhihui He, et~al. 2024.
\newblock Minicpm-v: A gpt-4v level mllm on your phone.
\newblock \emph{arXiv preprint arXiv:2408.01800}.

\bibitem[{Yin et~al.(2021)Yin, Li, Hu, Peng, and Chang}]{yin-etal-2021-broaden}
Da~Yin, Liunian~Harold Li, Ziniu Hu, Nanyun Peng, and Kai-Wei Chang. 2021.
\newblock \href {https://doi.org/10.18653/v1/2021.emnlp-main.162} {Broaden the vision: Geo-diverse visual commonsense reasoning}.
\newblock In \emph{Proceedings of the 2021 Conference on Empirical Methods in Natural Language Processing}, pages 2115--2129, Online and Punta Cana, Dominican Republic. Association for Computational Linguistics.

\bibitem[{Yu(2024)}]{yu2024han}
Qianqian Yu. 2024.
\newblock Han fu culture: the emphasis on chinese traditional cultural identity.
\newblock \emph{Trans/Form/A{\c{c}}{\~a}o}, 47(5):e02400157.

\bibitem[{Zhang(2024)}]{zhang2024from}
Jing Zhang. 2024.
\newblock \href {https://doi.org/10.1145/3689094.3689470} {From pixels to preservation: The power of large vision models in heritage content understanding}.
\newblock In \emph{Proceedings of the 6th Workshop on the AnalySis, Understanding and ProMotion of HeritAge Contents}, SUMAC '24, page 3–4, New York, NY, USA. Association for Computing Machinery.

\bibitem[{Zhou et~al.(2025)Zhou, Karidi, Liu, Garneau, Cao, Chen, Li, and Hershcovich}]{zhou-etal-2025-mapo}
Li~Zhou, Taelin Karidi, Wanlong Liu, Nicolas Garneau, Yong Cao, Wenyu Chen, Haizhou Li, and Daniel Hershcovich. 2025.
\newblock \href {https://aclanthology.org/2025.naacl-long.496/} {Does mapo tofu contain coffee? probing {LLM}s for food-related cultural knowledge}.
\newblock In \emph{Proceedings of the 2025 Conference of the Nations of the Americas Chapter of the Association for Computational Linguistics: Human Language Technologies (Volume 1: Long Papers)}, pages 9840--9867, Albuquerque, New Mexico. Association for Computational Linguistics.

\bibitem[{Zhu et~al.(2025)Zhu, Yu, Tong, and Hui}]{zhu2025exploring}
Zihao Zhu, Ao~Yu, Xin Tong, and Pan Hui. 2025.
\newblock Exploring llm-powered role and action-switching pedagogical agents for history education in virtual reality.
\newblock \emph{arXiv preprint arXiv:2505.02699}.

\bibitem[{Zou(2023)}]{zou2023research}
Qinling Zou. 2023.
\newblock Research on the effect of cross-cultural communication of chinese culture on youtube——evidence from hanfu.
\newblock \emph{Communications in Humanities Research}, 19:96--107.

\end{thebibliography}

\appendix

\section{Dataset details}\label{app:ann-instruction}


\subsection{Specific source of the dataset}
\label{app:data_source}

Online retailers at Taobao include:

\begin{itemize}
    \item \href{https://store.taobao.com/shop/view_shop.htm?appUid=RAzN8HWUJNgydcSuBxTz4JjzLN74hTZpqvSZLkkysvuo79RNYKv}{\begin{CJK}{UTF8}{gbsn}十三余小豆蔻儿\end{CJK}}
    \item \href{https://store.taobao.com/shop/view_shop.htm?appUid=RAzN8HWUyuiaB6ZeSe3VwW7jCSsh8tgadrRLy57JpgcDcEJVQRH}{\begin{CJK}{UTF8}{gbsn}织造司\end{CJK}}
    \item \href{https://store.taobao.com/shop/view_shop.htm?appUid=RAzN8HWRwutHBKaKfhDKF8oo89BT5MrUEd4gehi9RXfH3Uf9vfN}{\begin{CJK}{UTF8}{gbsn}花朝记旗舰店\end{CJK}}
    \item \href{https://store.taobao.com/shop/view_shop.htm?appUid=RAzN8HWNZWEdHpDqWmXibMt3sCGtCMgWJWhABGPEjDsyTJypG4H}{\begin{CJK}{UTF8}{gbsn}重回汉唐旗舰店\end{CJK}}
    \item \href{https://store.taobao.com/shop/view_shop.htm?appUid=RAzN8HAtjcLPXAqjX8aiArzpt4FnH}{\begin{CJK}{UTF8}{gbsn}山涧服饰\end{CJK}}
    \item \href{https://store.taobao.com/shop/view_shop.htm?appUid=RAzN8HWP8u45SayCsTBfu2pJcHcGZ9sf8GQiTjxrXxiRejRDeNw}{\begin{CJK}{UTF8}{gbsn}纨绮传统服饰\end{CJK}}
    \item \href{https://store.taobao.com/shop/view_shop.htm?appUid=RAzN8HWZg1cSSEQCVXnP31E4SKqsECy2YXvBncgW8wqbvUfsTN6}{\begin{CJK}{UTF8}{gbsn}洞庭汉风汉服男装\end{CJK}}
\end{itemize}

Chinese television series include:

\begin{itemize}
    \item \href{https://zh.m.wikipedia.org/wiki/大秦帝国之裂变}{\begin{CJK}{UTF8}{gbsn}大秦帝国 (The Qin Empirel)-Qin dynasty\end{CJK}}
    \item \href{https://zh.m.wikipedia.org/wiki/国色芳华}{\begin{CJK}{UTF8}{gbsn}国色芳华 (Flourished Peony)-Tang dynasty\end{CJK}}
    \item \href{https://zh.m.wikipedia.org/wiki/女医明妃传}{\begin{CJK}{UTF8}{gbsn}女医明妃传 (The Imperial Doctress)-Ming dynasty\end{CJK}}
    \item \href{https://zh.m.wikipedia.org/wiki/清平乐_(电视剧)}{\begin{CJK}{UTF8}{gbsn}清平乐 (Serenade of Peaceful Joy)-Song dynasty\end{CJK}}
    \item \href{https://baike.baidu.com/item/%E6%B0%B8%E5%AE%89%E6%A2%A6/60772328}{\begin{CJK}{UTF8}{gbsn}永安梦 (The Dream of Yong'an)-Tang dynasty\end{CJK}}
    \item \href{https://zh.m.wikipedia.org/wiki/长安十二时辰}{\begin{CJK}{UTF8}{gbsn}长安十二时辰 (The Longest Day In Chang'an)-Tang dynasty\end{CJK}}
    \item \href{https://zh.m.wikipedia.org/wiki/知否？知否？应是绿肥红瘦}{\begin{CJK}{UTF8}{gbsn}知否？知否？应是绿肥红瘦 (The Story of Ming Lan)-Song dynasty\end{CJK}}
\end{itemize}

\begin{figure*}[t]
    \centering
    \includegraphics[width=0.8\linewidth]{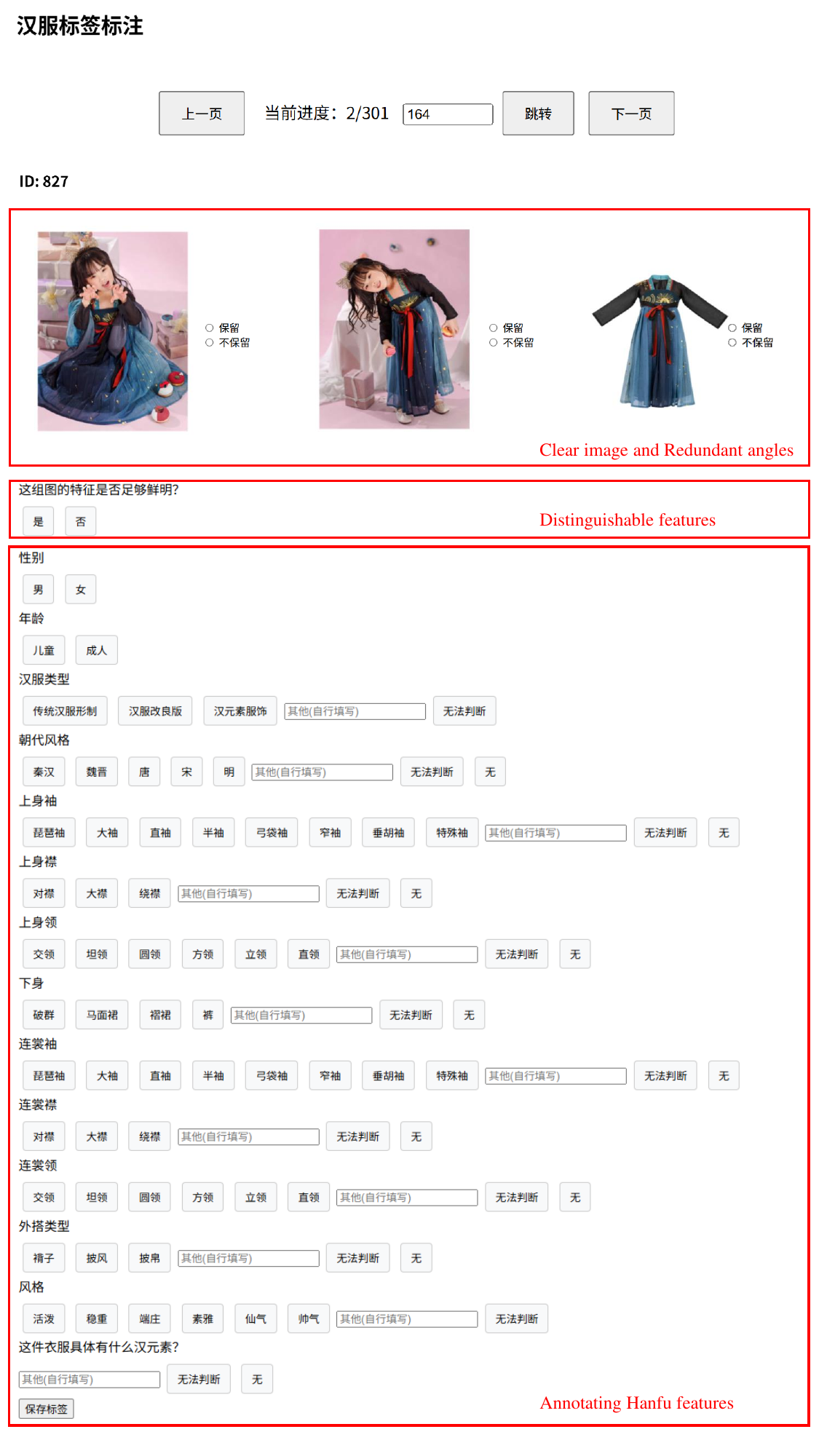}
    \caption{User interface of the annotation platform, where the annotation follows three steps regarding retaining/filtering images, assess whether the features of the Hanfu images are obvious, as well as fine-grained Hanfu feature annotation. }
    \label{fig:intervase}
\end{figure*}

\subsection{Annotation instruction}
\label{app:annotation_instruction}
The annotation process for Hanfu images involves the following steps:

1. \textbf{User Authentication}: Enter your personal name in pinyin at the username field and select the designated dataset assigned to you.

2. \textbf{Annotation Details}:
   \begin{enumerate}
       \item \textbf{Progress Tracking}: The "Progress" metric represents the ratio of completed pages to the total number of datasets. The "Current Progress" indicates the sequence of the current image within the dataset. To proceed to the next image, it is mandatory to complete all questions on the current page. The use of "Jump" is discouraged to prevent data transmission failure.
       \item \textbf{Image Preservation Decision}: Based on the image quality and content, determine whether the image can reflect the characteristics of the Hanfu style. If the image quality is low or the style features are extremely unclear, select "Not Retained".
       \item \textbf{Hanfu Category Definition}:
           \begin{itemize}
               \item \textbf{Traditional Hanfu style}: Refers to Hanfu that essentially conforms to the correct style with minimal modifications. Modern patterns are acceptable.
               \item \textbf{Improved Han clothing}: Refers to Hanfu that has a generally correct style but contains some modifications, or cases where individual items conform to the style but the overall combination does not align with traditional Hanfu (e.g., Mamian Qun + shirt, Sino-Western fusion).
               \item \textbf{Han-element clothing}: Refers to clothing with a modern overall silhouette that incorporates a very small number of style elements or features.
           \end{itemize}
       \item \textbf{Other Labels}: Refer to the "Various Hanfu Style Features" in the appendix for judgment. Note that if the Hanfu in the image is not a " lianchang", then " lianchang sleeve", " lianchang collar", and " lianchang lapel" should all be selected as "No". If uncertain, choose "Unable to Judge".
       \item \textbf{Specific Hanfu Elements Identification}: For the question "What specific Hanfu elements does this piece of clothing have?", consider the clothes appearing on the current page. Despite potential slight differences in style and color, focus on several more obvious and distinctive Hanfu elements.
       \item \textbf{Data Preservation}: After completing all questions on the page, it is imperative to click "Save Labels" before proceeding to the next page to avoid data transmission issues.
   \end{enumerate}

An appendix containing illustrations of various Hanfu style features is provided for reference.

\subsection{Annotation platform}
\label{app:annotation_interface}
As shown in the Figure~\ref{fig:intervase}, there are three main functions of the annotation window: first, the annotator needs to filter out clear and non-repeating high-quality images; Second, the annotator needs to distinguish whether the Hanfu features in these Hanfu pictures are distinct enough; Three: Label the characteristics of these Hanfu.

\subsection{Dataset statistics}
\label{app:dataset_statistics}
\begin{figure*}[t]
    \centering
    \includegraphics[width=1\linewidth]{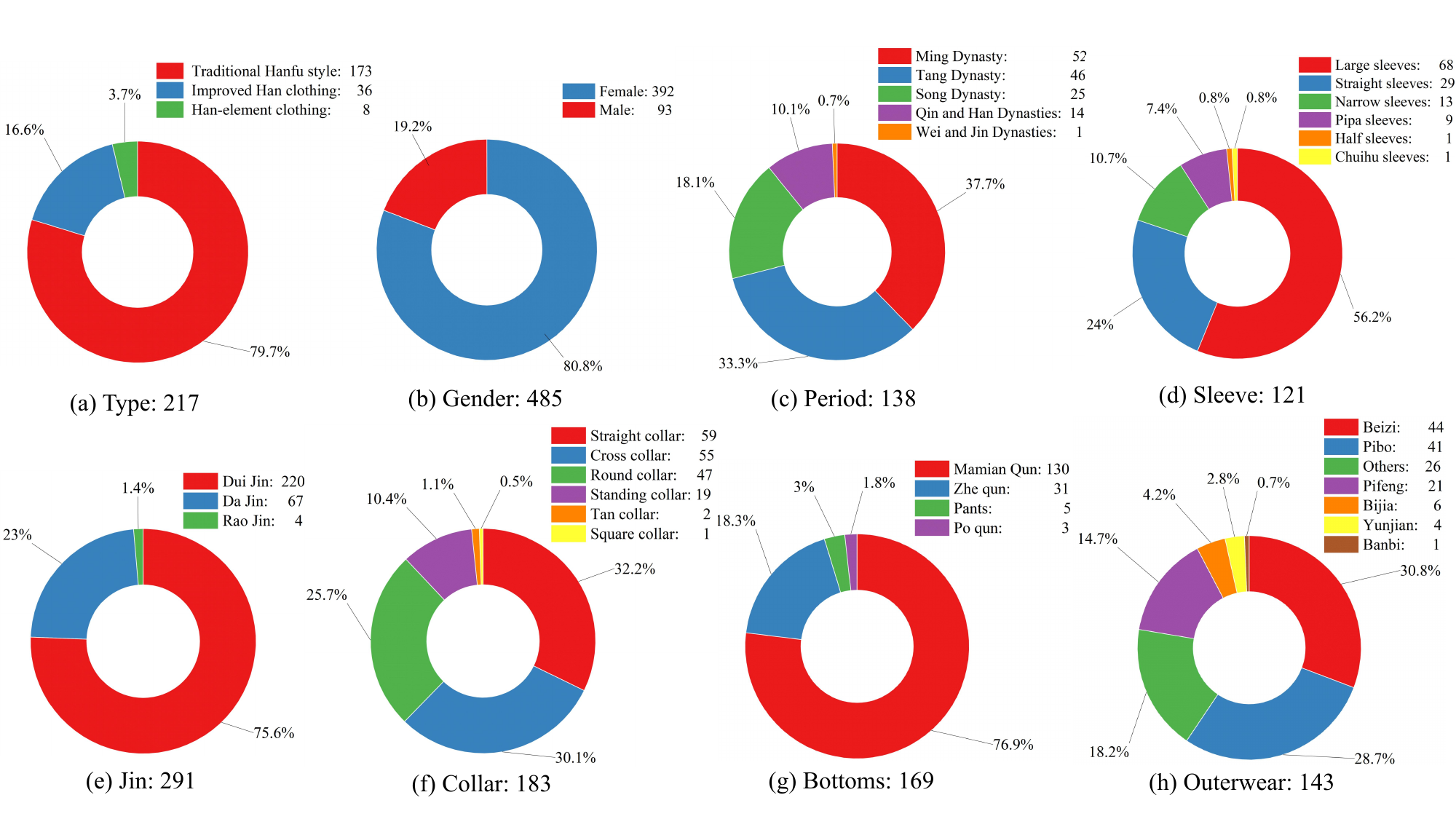}
    \caption{Statistics and distribution of the collected temporal-culture feature annotations.}
    \label{fig:dataset_stat}
\end{figure*}

The dataset statistics on feature distribution are presented in the Figure~\ref{fig:dataset_stat}.


\begin{figure*}[ht]
    \centering
    \includegraphics[width=1.0\linewidth]{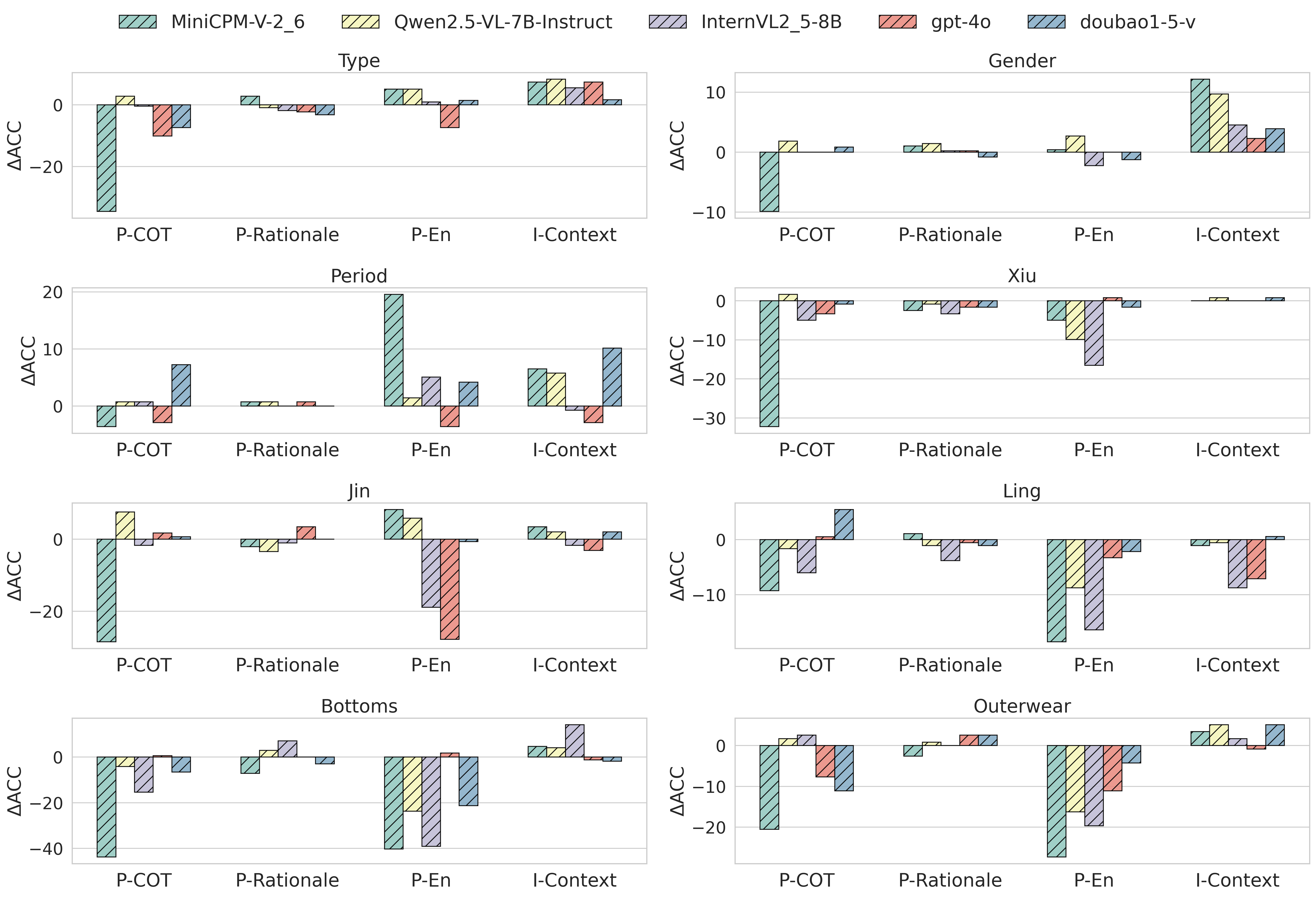}
    \caption{\textbf{SVQA}: Fine-grained performance variations of VLMs when applying different prompts and including image contexts compared to the base Chinese prompt. \textit{P-COT}: step-by-step reasoning instructions; \textit{P-Rationale}: requiring answer rationales; \textit{P-En}: English prompts; \textit{I-context}: images with background retained.}
    \label{fig:svqa_compare}
\end{figure*}

\begin{figure*}[ht]
    \centering
    \includegraphics[width=1.0\linewidth]{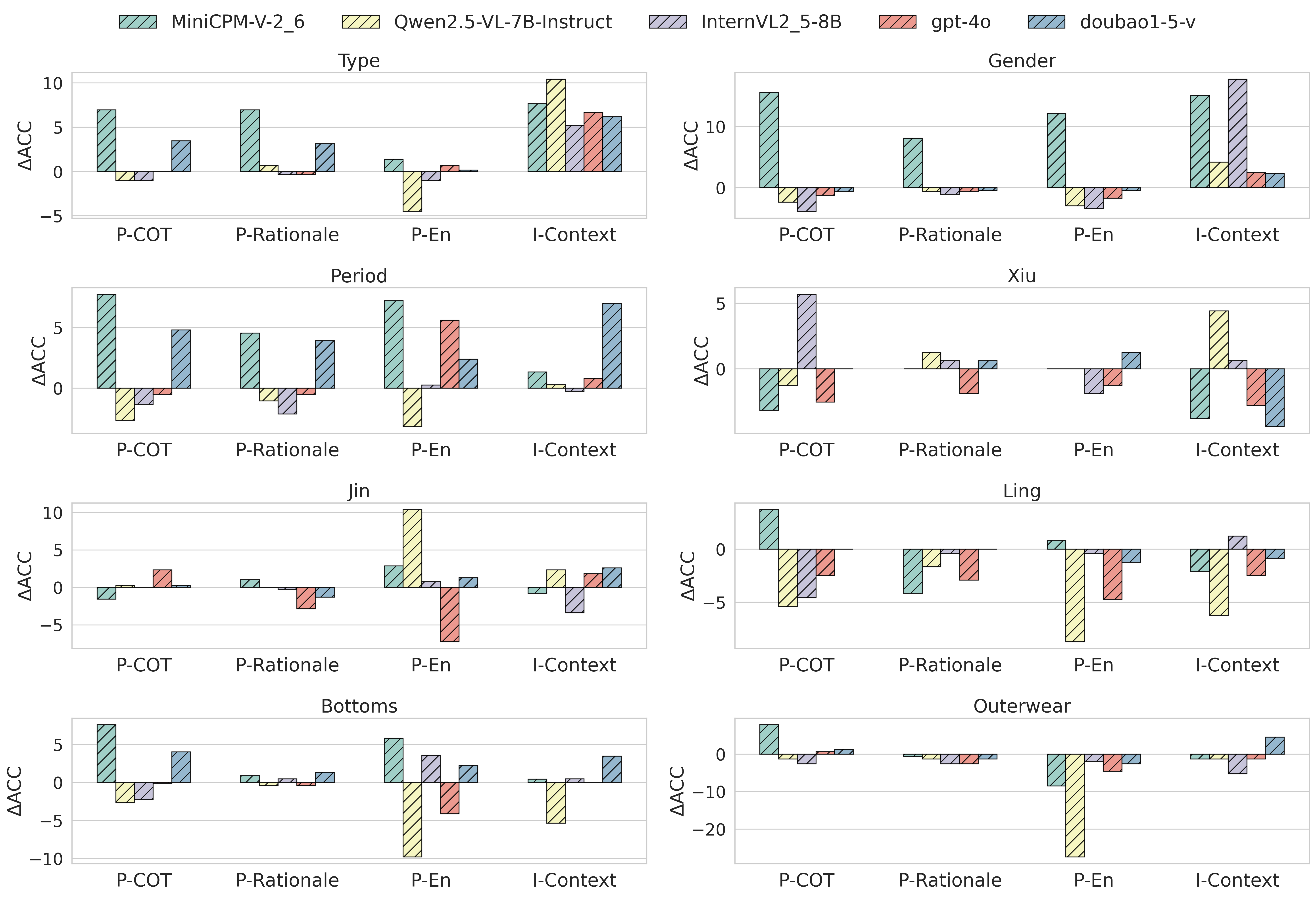}
    \caption{\textbf{MVQA}: Fine-grained Performance changes of VLMs with different prompt types and image contexts compared to the base Chinese prompt. \textit{P-COT}: step-by-step reasoning instructions; \textit{P-Rationale}: requiring answer rationales; \textit{P-En}: English prompts; \textit{I-context}: images with background retained.}
    \label{fig:mvqa_compare}
\end{figure*}

\subsection{Base question templates and feature value}\label{app:question templates}
Table~\ref{tab:value} lists the specific values for all meta-information. Table~\ref{tab:svqa-question} provides the base questions for the SVQA task, and Table~\ref{tab:MVQA} outlines the base questions for the MVQA task.

\begin{table*}[t]
\resizebox{\textwidth}{!}{
\begin{tabular}{@{}lp{15cm}@{}}
\toprule
\textbf{Category} & \textbf{Elements}                                                                                                              \\ \midrule

Type      & \begin{CJK}{UTF8}{gbsn}传统汉服形制 Traditional Hanfu style,  汉服改良版 Improved Han clothing, 汉元素服饰 Han-element clothing                        \end{CJK}\\
Gender    &\begin{CJK}{UTF8}{gbsn} 男 Male, 女 Female                                                                                                \end{CJK}\\
Period    & \begin{CJK}{UTF8}{gbsn}秦汉时期 Qin and Han Dynasties, 魏晋时期 Wei and Jin Dynasties, 唐朝 Tang Dynasty, 宋朝 Song Dynasty. 明朝 Ming Dynasty      \end{CJK} \\
Sleeve       & \begin{CJK}{UTF8}{gbsn}窄袖 Narrow sleeves, 直袖 Straight sleeves, 半袖 Half sleeves, 琵琶袖 Pipa sleeves, 垂胡袖 Chuihu sleeves, 大袖 Large sleeves\end{CJK} \\
Jin       & \begin{CJK}{UTF8}{gbsn}大襟 Da Jin, 对襟 Dui Jin, 绕襟 Rao Jin                                                                              \end{CJK} \\
Collar      & \begin{CJK}{UTF8}{gbsn}直领 Straight collar, 坦领 Tan collar, 圆领 Round collar, 方领 Square collar, 立领 Standing collar, 交领 Cross collar       \end{CJK}\\
Bottoms   &\begin{CJK}{UTF8}{gbsn} 破群Po qun, 裤 Pants, 马面裙 Mamian qun, 褶裙 Zhe qun                                                                  \end{CJK} \\
Outerwear &\begin{CJK}{UTF8}{gbsn} 比甲 Bijia, 半臂 Banbi, 云肩 Yunjian, 褙子 Beizi, 披帛 Pibo, 披风 Pifeng                                                    \end{CJK}\\ \bottomrule
\end{tabular}}
\caption{Feature value list.}
\label{tab:svqa-question}
\label{tab:value}
\end{table*}

\begin{table*}[t]
\resizebox{\textwidth}{!}{
\begin{tabular}{@{}lp{15cm}@{}}
\toprule
\textbf{Meta info} & \textbf{Question}                                                                                                              \\ \midrule
Type               & \begin{CJK}{UTF8}{gbsn}图片中的服饰通常属于以下哪个类型？\end{CJK}Which type does the clothing in this image typically belong to, among the following options?                  \\
Gender             & \begin{CJK}{UTF8}{gbsn}图片中的服饰通常适合什么性别？\end{CJK}Which gender is the clothing in this image typically suitable for, among the following options?                 \\
Period             & \begin{CJK}{UTF8}{gbsn}图片中的服饰属于以下哪个时期的风格？\end{CJK}Which historical period does the style of the clothing in this image belong to, among the following options? \\
Sleeve             & \begin{CJK}{UTF8}{gbsn}图片中服饰的袖子属于以下哪种类型？\end{CJK}Which type does the sleeves of the clothing in this image belong to, among the following options?             \\
Jin                & \begin{CJK}{UTF8}{gbsn}图片中服饰的襟型属于以下哪种类型？\end{CJK}Which type does the lapel style of the clothing in this image belong to, among the following options?         \\
Collar             & \begin{CJK}{UTF8}{gbsn}图片中服饰的领型属于以下哪种类型？\end{CJK}Which type does the collar style of the clothing in this image belong to, among the following options?        \\
Bottoms            & \begin{CJK}{UTF8}{gbsn}图片中服饰的下身是什么类型的？\end{CJK}Which type does the lower garment of the clothing in this image belong to, among the following options?         \\
Outerwear          & \begin{CJK}{UTF8}{gbsn}图片中服饰的外搭是什么？\end{CJK}Which type does the outer layer of the clothing in this image belong to, among the following options?              \\ \bottomrule
\end{tabular}}
\caption{Base questions in SVQA task.}
\label{tab:svqa-question}
\end{table*}


\begin{table*}[]
\resizebox{0.95\textwidth}{!}{
\begin{tabular}{@{}p{1.1cm}p{22cm}@{}}
\toprule
    Meta                   & Question                                                                                                                                                                                     \\ \midrule
\multirow{7}{*}{Type}         & \begin{CJK}{UTF8}{gbsn}以下图片中的服饰属于汉服改良版的是？Which of the following pictures of clothing belongs to the Improved Han   clothing?\end{CJK}\\
                              & \begin{CJK}{UTF8}{gbsn}以下图片中的服饰属于传统汉服形制的是？Which of the following images represents a traditional Hanfu style?\end{CJK}  \\
                              & \begin{CJK}{UTF8}{gbsn}以下图片中不属于汉服改良版的服饰是？Which of the following pictures of clothing does not belong to the improved Han clothing?\end{CJK}  \\
                              & \begin{CJK}{UTF8}{gbsn}以下图片中不属于传统汉服形制的服饰是？ Which of the   following images does NOT belong to traditional Hanfu styles?\end{CJK}\\
                              & \begin{CJK}{UTF8}{gbsn}以下图片中的服饰属于汉元素服饰的是？Which of the following images represents Han element clothing?\end{CJK}  \\
                              & \begin{CJK}{UTF8}{gbsn}以下图片中形制与其他服饰不同的是？   Which of the following images shows a different structural style compared to   the other garments?\end{CJK} \\
                              & \begin{CJK}{UTF8}{gbsn}以下图片中风格与其他服饰不同的有？Which of the following images shows a different overall style compared to the   others?\end{CJK} \\ \midrule
\multirow{5}{*}{Gender}       & \begin{CJK}{UTF8}{gbsn}以下图片中的服饰属于男性服饰的是？ Which of the following pictures of clothing belongs to male's clothing? \end{CJK}  \\
                              & \begin{CJK}{UTF8}{gbsn}以下图片中不属于男性服饰的是？Which of the following images does NOT depict male attire? \end{CJK}                                                                        \\
                              & \begin{CJK}{UTF8}{gbsn}以下图片中的服饰属于女性服饰的是？Which of the following images depicts female's Hanfu attire? \end{CJK}                                                                    \\
                              & \begin{CJK}{UTF8}{gbsn}以下图片中不属于女性服饰的是？Which of the following images does NOT show women's attire? \end{CJK}                                                                       \\
                              & \begin{CJK}{UTF8}{gbsn}以下图片中服饰性别属性与其他图片不同的是？Which of the following images shows different gender attributes compared to   the others?\end{CJK}  \\ \midrule
\multirow{12}{*}{Period}      & \begin{CJK}{UTF8}{gbsn}以下图片中属于唐朝以后时期的服饰有？ Which of the following images shows clothing styles from after the Tang Dynasty?\end{CJK}                                                \\
                              & \begin{CJK}{UTF8}{gbsn}以下图片中属于魏晋时期以后时期的服饰有？ Which of the following images shows clothing styles from after the Wei-Jin   period? \end{CJK}  \\
                              & \begin{CJK}{UTF8}{gbsn}以下的服装中比以上图片中的服饰更古老的有？Which of the following garments is older than those shown in the images   above? \end{CJK}  \\
                              & \begin{CJK}{UTF8}{gbsn}以下图片中的服饰属于秦汉时期的风格的有？Which of the following images represents Qin and Han Dynasties' style   clothing?\end{CJK} \\
                              & \begin{CJK}{UTF8}{gbsn}以下图片中的服饰属于明朝的风格的有？Which of the following images represents Ming Dynasty-style clothing?   \end{CJK}                                                        \\
                              & \begin{CJK}{UTF8}{gbsn}以下图片中不属于唐朝的服饰有？Which of the following images does NOT show Tang Dynasty attire? \end{CJK}                \\
                              & \begin{CJK}{UTF8}{gbsn}以下图片中不属于宋朝的服饰有？Which of the following images does NOT represent Song Dynasty-era clothing?   \end{CJK}   \\
                              & \begin{CJK}{UTF8}{gbsn}以下图片中不属于明朝的服饰有？Which of the following images does NOT show Ming Dynasty-era attire?   \end{CJK}                                                            \\
                              & \begin{CJK}{UTF8}{gbsn}以下图片中的服饰属于魏晋时期的风格的有？   Which of the following images represents Wei-Jin period style   clothing?\end{CJK}                                                     \\
                              & \begin{CJK}{UTF8}{gbsn}以下图片中的服饰属于唐朝的风格的有？   Which of the following images represents Tang Dynasty-style   clothing?\end{CJK}                                                         \\
                              & \begin{CJK}{UTF8}{gbsn}以下图片中属于宋朝以后时期的服饰有？   Which of the following images displays clothing styles from after the Song   Dynasty? \end{CJK}                                          \\
                              & \begin{CJK}{UTF8}{gbsn}以下图片中的服饰属于宋朝的风格的有？   Which of the following images represents Song Dynasty-style   clothing?\end{CJK}                                                         \\ \midrule
\multirow{11}{*}{Sleeve}         & \begin{CJK}{UTF8}{gbsn}以下图片中服饰袖型与其他图片不同的是？ Which of the following images   has a sleeve style that differs from the others?\end{CJK}                                                 \\
                              & \begin{CJK}{UTF8}{gbsn}以下图片中服饰袖型属于垂胡袖的是？   Which of the following images features the 'Chuihu sleeve'? \end{CJK}                                                                     \\
                              & \begin{CJK}{UTF8}{gbsn}以下图片中服饰袖型不属于琵琶袖的是？Which   of the following images does NOT feature pipa sleeves? \end{CJK}                                                                    \\
                              & \begin{CJK}{UTF8}{gbsn}以下图片中服饰袖型属于直袖的是？   Which of the following images displays straight sleeves?\end{CJK}                                                                          \\
                              & \begin{CJK}{UTF8}{gbsn}以下图片中服饰袖型属于窄袖的是？   Which of the following images features narrow sleeves?\end{CJK}                                                                            \\
                              & \begin{CJK}{UTF8}{gbsn}以下图片中服饰袖型不属于直袖的是？   Which of the following images does NOT have straight sleeves?\end{CJK}                                                                    \\
                              & \begin{CJK}{UTF8}{gbsn}以下图片中服饰袖型属于琵琶袖的是？   Which of the following images displays the 'Pipa sleeve'? \end{CJK}                                                                       \\
                              & \begin{CJK}{UTF8}{gbsn}以下图片中服饰袖型不属于窄袖的是？   Which of the following images does NOT have narrow sleeves? \end{CJK}                                                                     \\
                              & \begin{CJK}{UTF8}{gbsn}以下图片中服饰袖型不属于大袖的是？   Which of the following images does NOT feature large sleeves? \end{CJK}                                                                   \\
                              & \begin{CJK}{UTF8}{gbsn}以下图片中服饰袖型属于半袖的是？   Which of the following images features half sleeves? \end{CJK}                                                                             \\
                              & \begin{CJK}{UTF8}{gbsn}以下图片中服饰袖型属于大袖的是？   Which of the following images shows large sleeves? \end{CJK}                                                                               \\ \midrule
\multirow{11}{*}{Collar}        & \begin{CJK}{UTF8}{gbsn}以下图片中服饰领型属于交领的是？ Which of the following images   shows the 'Cross collar' neckline style?\end{CJK}                                                            \\
                              & \begin{CJK}{UTF8}{gbsn}以下图片中服饰领型不属于直领的是？   Which of the following images does NOT display the 'straight collar'   neckline?\end{CJK}                                                 \\
                              & \begin{CJK}{UTF8}{gbsn}以下图片中服饰领型属于直领的是？   Which of the following images shows a straight collar neckline?\end{CJK}                                                                   \\
                              & \begin{CJK}{UTF8}{gbsn}以下图片中服饰领型与其他图片不同的是？   Which of the following images has a neckline style that differs from the   others?\end{CJK}                                             \\
                              & \begin{CJK}{UTF8}{gbsn}以下图片中服饰领型属于立领的是？   Which of the following images has a standing collar?\end{CJK}                                                                              \\
                              & \begin{CJK}{UTF8}{gbsn}以下图片中服饰领型属于圆领的是？   Which of the following images features a round collar neckline?\end{CJK}                                                                   \\
                              & \begin{CJK}{UTF8}{gbsn}以下图片中服饰领型不属于立领的是？   Which of the following images does NOT have a standing collar?\end{CJK}                                                                   \\
                              & \begin{CJK}{UTF8}{gbsn}以下图片中服饰领型属于方领的是？   Which of the following images features a square collar   neckline?\end{CJK}                                                                \\
                              & \begin{CJK}{UTF8}{gbsn}以下图片中服饰领型属于坦领的是？   Which of the following images has a 'Tan collar'?\end{CJK}                                                                                 \\
                              & \begin{CJK}{UTF8}{gbsn}以下图片中服饰领型不属于圆领的是？   Which of the following images does NOT have a round collar? \end{CJK}                                                                     \\
                              & \begin{CJK}{UTF8}{gbsn}以下图片中服饰领型不属于交领的是？   Which of the following images does NOT have a cross-collar? \end{CJK}                                                                     \\ \midrule
\multirow{6}{*}{Jin}          & \begin{CJK}{UTF8}{gbsn}以下图片中服饰襟型与其他图片不同的是？ Which of the following images   shows a different garment closure type compared to the others?\end{CJK}                                   \\
                              & \begin{CJK}{UTF8}{gbsn}以下图片中服饰襟型不属于大襟的是？   Which of the following images does NOT feature the 'Dajin' closure style?   \end{CJK}                                                     \\
                              & \begin{CJK}{UTF8}{gbsn}以下图片中服饰襟型不属于对襟的是？   Which of the following images does NOT show the 'Duijin' closure style?   \end{CJK}                                                       \\
                              & \begin{CJK}{UTF8}{gbsn}以下图片中服饰襟型属于绕襟的是？   Which of the following images features a wrapped closure ('Raojin') style?   \end{CJK}                                                     \\
                              & \begin{CJK}{UTF8}{gbsn}以下图片中服饰襟型属于对襟的是？   Which of the following images shows a symmetrical front closure ('Duijin')?   \end{CJK}                                                    \\
                              & \begin{CJK}{UTF8}{gbsn}以下图片中服饰襟型属于大襟的是？   Which of the following images shows the 'Dajin'?\end{CJK}                                                                                  \\ \midrule
\multirow{7}{*}{Bottoms}      & \begin{CJK}{UTF8}{gbsn}以下图片中下半身服饰种类属于马面裙的是？ Among the following pictures   which type of lower body clothing belongs to 'Mamian Qun?'\end{CJK}                                       \\
                              & \begin{CJK}{UTF8}{gbsn}以下图片中下半身服饰种类属于裤的是？   Which of the following images features lower-body clothing categorized as   trousers? \end{CJK}                                          \\
                              & \begin{CJK}{UTF8}{gbsn}以下图片中下半身服饰种类不属于马面裙的是？   Which of the following images does NOT depict a 'Mamian qun' as the   lower-body attire?  \end{CJK}                                   \\
                              & \begin{CJK}{UTF8}{gbsn}以下图片中下半身服饰种类与其他图片不同的是？   Which of the following images shows a different lower-body garment style   compared to the others?  \end{CJK}                        \\
                              & \begin{CJK}{UTF8}{gbsn}以下图片中下半身服饰种类属于破群的是？   Which of the following images depicts a 'Poqun' as the lower-body   garment?  \end{CJK}                                                 \\
                              & \begin{CJK}{UTF8}{gbsn}以下图片中下半身服饰种类属于褶裙的是？   Which of the following images depicts pleated skirts ('Zhequn')?  \end{CJK}                                                             \\
                              & \begin{CJK}{UTF8}{gbsn}以下图片中下半身服饰种类不属于褶裙的是？   Which of the following images does NOT depict pleated skirts ('Zhequn')?   \end{CJK}                                                   \\ \midrule
\multirow{12}{*}{Outerwear}   & \begin{CJK}{UTF8}{gbsn}以下图片中外搭服饰种类属于披帛的是？ Which of the following types of   outerwear in the pictures belongs to the category of 'Pibo'? \end{CJK}                                   \\
                              & \begin{CJK}{UTF8}{gbsn}以下图片中外搭服饰种类不属于披帛的是？   Which of the following images shows an outer garment that is NOT a   'Pibo'?  \end{CJK}                                                 \\
                              & \begin{CJK}{UTF8}{gbsn}以下图片中外搭服饰种类不属于褙子的是？   Which of the following images features an outer garment that is NOT a   'Beizi'?  \end{CJK}                                             \\
                              & \begin{CJK}{UTF8}{gbsn}以下图片中外搭服饰种类与其他图片不同的是？   Which of the following images features an outer garment that differs from the   others?  \end{CJK}                                    \\
                              & \begin{CJK}{UTF8}{gbsn}以下图片中外搭服饰种类不属于云肩的是？   Which of the following images does NOT include a 'Yunjian'? \end{CJK}                                                                   \\
                              & \begin{CJK}{UTF8}{gbsn}以下图片中外搭服饰种类不属于披风的是？   Which of the following images does NOT include a 'Pifeng' as outerwear?   \end{CJK}                                                     \\
                              & \begin{CJK}{UTF8}{gbsn}以下图片中外搭服饰种类属于比甲的是？   Which of the following images includes a 'Bijia' as outerwear?  \end{CJK}                                                                \\
                              & \begin{CJK}{UTF8}{gbsn}以下图片中外搭服饰种类属于云肩的是？   Which of the following images features a 'Yunjian'?  \end{CJK}                                                                           \\
                              & \begin{CJK}{UTF8}{gbsn}以下图片中外搭服饰种类属于披风的是？   Which of the following images includes a 'Pifeng'? \end{CJK}                                                                             \\
                              & \begin{CJK}{UTF8}{gbsn}以下图片中外搭服饰种类属于半臂的是？   Which of the following images includes a 'Banbi' as outerwear?  \end{CJK}                                                                \\
                              & \begin{CJK}{UTF8}{gbsn}以下图片中外搭服饰种类属于褙子的是？   Which of the following images features a 'Beizi'?  \end{CJK}                                                                             \\
                              & \begin{CJK}{UTF8}{gbsn}以下图片中外搭服饰种类不属于比甲的是？   Which of the following images does NOT include a 'Bijia'? \end{CJK}                                                                     \\ \bottomrule

    \end{tabular}}
    \caption{Base questions in MVQA task.}
    \label{tab:MVQA}
\end{table*}

\section{Prompts for VQA}
\label{app:vqa_prompt}
The prompts involved in SVQA and MVQA are shown in Table~\ref{tab:prompt_svqa} and Table~\ref{tab:prompt_mvqa}.

\begin{CJK}{UTF8}{gbsn}
\begin{table*}[ht]
    \centering
    \resizebox{\textwidth}{!}{
    \begin{tabular}{|p{20.5cm}|}
\hline
\textbf{Prompt 1} \\
\hline
你是一位了解中国汉服的专家。我将向你提供一张服饰图片，并附带一个问题和几个选项。请仔细观察图片，依据服饰的特点，选择最符合图片信息的答案。
\\
 \#\# 输出要求：\\
请以 JSON 格式输出答案，包含以下字段：\\
"答案"：选择的正确选项（如 "A"、"B"、"C" 或 "D"）。\\
\#\# 输入
\\
 \hline
\textbf{Prompt 2} \\
\hline
你是一名汉服领域的专家，熟悉传统服饰的款式与特点。我将提供一张服饰图片，并附上一个问题及几个选项。请依据图片中的服饰特征，选择最合适的答案。
\\
 \#\# 输出要求：\\
请以 JSON 格式输出答案，包含以下字段：\\
"答案"：选择的正确选项（如 "A"、"B"、"C" 或 "D"）。\\
\#\# 输入
\\
 \hline
\textbf{Prompt 3} \\
\hline
你是一位对传统服饰文化及其现代改良有深入了解的专家。我将提供一张服饰图片，并附带一个问题和几个选项。请依据图片中的服饰特点，选择最符合图片信息的答案。
\\
 \#\# 输出要求：\\
请以 JSON 格式输出答案，包含以下字段：\\
"答案"：选择的正确选项（如 "A"、"B"、"C" 或 "D"）。\\
\#\# 输入
\\
 \hline
\textbf{Prompt 4} \\
\hline
你是一位对中国传统服饰有深入了解的专家。我将提供一张服饰图片，并附带一个问题和几个选项。请依据图片中的服饰特点，选择最符合图片信息的答案。
\\
 \#\# 输出要求：\\
请以 JSON 格式输出答案，包含以下字段：\\
"答案"：选择的正确选项（如 "A"、"B"、"C" 或 "D"）。\\
\#\# 输入
\\
 \hline
\textbf{Prompt 5} \\
\hline
你是一位对中国传统服饰——汉服，有着深厚研究和丰富知识的专家。我将提供一张服饰图片，并附带一个问题和几个选项。请依据图片中的服饰特点，选择最符合图片信息的答案。
\\
 \#\# 输出要求：\\
请以 JSON 格式输出答案，包含以下字段：\\
"答案"：选择的正确选项（如 "A"、"B"、"C" 或 "D"）。\\
\#\# 输入
\\
 \hline

\textbf{P-COT} \\
\hline
你是一位汉服研究专家，我将提供一张服饰图片并附带一个问题，请按照以下要求完成任务：\\
1. 观察图片，分析服饰的具体特征（例如形制、领型、襟型等）。\\
2. 根据这些特征，与问题中的选项逐一匹配，排除不符合的选项。\\
3. 得出最符合图片信息的最终答案，并详细说明你的推理过程。\\
\\
 \#\#输出要求：\\
请以 JSON 格式输出答案，包含以下字段：\\
"答案"：选择的正确选项（如 "A"、"B"、"C" 或 "D"）。\\
"推理"：详细描述你获取答案的推理过程。\\
\#\# 输入
\\
 \hline

\textbf{P-Rationale} \\
\hline
你是一位了解中国汉服的专家。我将向你提供一张服饰图片，并附带一个问题和几个选项。你的任务是根据图片中服饰的特点，选择最符合图片信息的正确答案，并详细解释你的选择理由。
\\
 \#\#输出要求：\\
请以 JSON 格式输出答案，包含以下字段：\\
"答案"：选择的正确选项（如 "A"、"B"、"C" 或 "D"）。\\
"原因"：详细解释选择该答案的理由，结合图片的细节和汉服的特点。
\\
\#\# 输入
\\
 \hline

\textbf{P-En} \\
\hline
You are an expert in traditional Chinese Hanfu. I will provide you with an image of clothing along with a question and several options. Please carefully observe the image and, based on the characteristics of the clothing, select the answer that best matches the information in the image.
\\
 \#\#Output Requirements:\\
Please output the answer in JSON format, including the following fields:\\
"answer": The selected correct option (e.g., "A", "B", "C", or "D").\\

\\
\#\# Input
\\
 \hline

    \end{tabular}}
    \caption{Different prompt templates for SVQA.}
    \label{tab:prompt_svqa}
\end{table*}
\end{CJK}
\begin{CJK}{UTF8}{gbsn}
\begin{table*}[ht]
    \centering
    \resizebox{\textwidth}{!}{
    \begin{tabular}{|p{20.5cm}|}
\hline
\textbf{Prompt 1} \\
\hline
你是一位了解中国汉服的专家。请根据提供的四张服饰图片（分别对应选项A、选项B、选项C、选项D），回答以下问题。从给定的选项中选择一个最符合的答案。
\\
 \#\# 输出要求：\\
请以 JSON 格式输出答案，包含以下字段：\\
"答案"：选择的正确选项（如 "A"、"B"、"C" 或 "D"）。\\
\#\# 输入
\\
 \hline
\textbf{Prompt 2} \\
\hline
你是一名汉服领域的专家，熟悉传统服饰的款式与特点。我将给出一个问题，请根据提供的四张汉服图片（分别对应选项A、选项B、选项C、选项D），选择最符合问题要求的答案。\\
 \#\# 输出要求：\\
请以 JSON 格式输出答案，包含以下字段：\\
"答案"：选择的正确选项（如 "A"、"B"、"C" 或 "D"）。\\
\#\# 输入
\\
 \hline
\textbf{Prompt 3} \\
\hline
你是一位对传统服饰文化及其现代改良有深入了解的专家。我将给出一个问题，以及四张服饰图片作为选项，请选择最符合问题要求的答案。\\
 \#\# 输出要求：\\
请以 JSON 格式输出答案，包含以下字段：\\
"答案"：选择的正确选项（如 "A"、"B"、"C" 或 "D"）。\\
\#\# 输入
\\
 \hline
\textbf{Prompt 4} \\
\hline
你是一位对中国传统服饰有深入了解的专家。我将给出一个问题，以及四个选项，每个选项为一张服饰图片，请选择最符合问题要求的答案。\\
 \#\# 输出要求：\\
请以 JSON 格式输出答案，包含以下字段：\\
"答案"：选择的正确选项（如 "A"、"B"、"C" 或 "D"）。\\
\#\# 输入
\\
 \hline
\textbf{Prompt 5} \\
\hline
你是一位对中国传统服饰——汉服，有着深厚研究和丰富知识的专家。请根据提供的四张服饰图片（分别对应选项A、选项B、选项C、选项D），回答一下问题，选择最符合问题要求的一个答案。\\
 \#\# 输出要求：\\
请以 JSON 格式输出答案，包含以下字段：\\
"答案"：选择的正确选项（如 "A"、"B"、"C" 或 "D"）。\\
\#\# 输入
\\
 \hline

\textbf{P-COT} \\
\hline
你是一位了解中国汉服的专家。请根据提供的四张服饰图片（分别对应选项A、选项B、选项C、选项D），请按照以下要求完成任务：\\
1. 观察每张图片，分析每张图片中服饰的具体特征（例如形制、领型、襟型等）。\\
2. 根据这些特征，将每张图片与问题逐一匹配，排除不符合的选项。\\
3. 从给定的选项中选择一个最符合的答案，并详细说明你的推理过程。\\
\\
 \#\#输出要求：\\
请以 JSON 格式输出答案，包含以下字段：\\
"答案"：选择的正确选项（如 "A"、"B"、"C" 或 "D"）。\\
"推理"：详细描述你获取答案的推理过程。\\
\#\# 输入
\\
 \hline

\textbf{P-Rationale} \\
\hline
你是一位了解中国汉服的专家。请根据提供的四张服饰图片（分别对应选项A、选项B、选项C、选项D），回答以下问题。从给定的选项中选择一个最符合的答案，并详细解释你的选择理由。\\
 \#\#输出要求：\\
请以 JSON 格式输出答案，包含以下字段：\\
"答案"：选择的正确选项（如 "A"、"B"、"C" 或 "D"）。\\
"原因"：详细解释选择该答案的理由。
\\
\#\# 输入
\\
 \hline

\textbf{P-En} \\
\hline
You are an expert in traditional Chinese Hanfu. Please answer the following question based on the four provided clothing images (labeled Option A, Option B, Option C, and Option D). Select the most appropriate answer from the given options.
\\
 \#\#Output Requirements:\\
Please output the answer in JSON format, including the following fields:\\
"answer": The selected correct option (e.g., "A", "B", "C", or "D").\\

\\
\#\# Input
\\
 \hline

    \end{tabular}}
    \caption{Different prompt templates used for MVQA.}
    \label{tab:prompt_mvqa}
\end{table*}
\end{CJK}

\section{Task 2: Cultural Image Transcreation}


\subsection{Caption Extraction and Caption Edit Prompts}\label{app:Quantitative Metrics of Task2}

\begin{CJK}{UTF8}{gbsn}
\begin{table*}[ht]
    \centering
    \resizebox{\textwidth}{!}{
    \begin{tabular}{|p{20.5cm}|}
    \hline
\textbf{Caption Extraction Prompt} \\
\hline
提供这件传统中国汉服服装的详细描述，包括其设计元素、颜色、图案和特色
\\
 Provide a   detailed description of this traditional Chinese Hanfu clothing, including   its design elements, colors, patterns, and features.\\

 \hline
\textbf{Caption Edit Prompt} \\
\hline
Transform this traditional Hanfu description: "\{caption\}" into   a modern clothing design by strategically incorporating its key elements.   Your task is to: 1. Choose ONE specific modern garment type as the base   (select from: hoodie, blazer, casual wear, sportswear, trench coat,   streetwear, or business attire) 2. Identify 3-5 distinctive elements from the   original Hanfu description (such as collar style, sleeve design, waist   details, fabric patterns, or color schemes) 3. Describe exactly how these   Hanfu elements are integrated into the modern garment: - WHERE each element   is placed on the modern garment - HOW each element is adapted to suit   contemporary fashion - WHY these particular elements were chosen (cultural   significance) 4. Ensure the final design: - Is primarily a modern, wearable   garment for everyday contexts - Clearly displays its Hanfu inspiration   through intentional design choices - Balances contemporary style with   traditional Chinese aesthetics - Appeals to modern fashion sensibilities   while honoring cultural heritage The output should provide a detailed   description of this hybrid garment, suitable for image generation, focusing   on both visual appearance and construction details.
\\
 \hline

    \end{tabular}}
    \caption{Prompts used in image caption and caption-edit stages in cultural image transcreation task. }
    \label{tab:task2_prompt}
\end{table*}
\end{CJK}
Table~\ref{tab:task2_prompt} presents the specific prompts used in cultural image transcreation's VLM caption extraction and caption edit processes.

\subsection{Quantitative Metric}\label{app:Quantitative Metrics of Task2}
Table~\ref{tab:human_detail} presents the six quantitative evaluation metrics employed in our study. Below is a detailed introduction to the metrics:

C0: The visual-change metric was established to assess whether the model-generated images exhibit effective modifications relative to the input images. This metric evaluates the extent of image editing performed by the model, with a score of 1 (negligible visual changes), 3 (moderate visual changes), and 5 (significant visual changes).

C1: The semantic-equivalence metric was designed to verify whether the modified images retain correct semantic information. It assesses the model's ability to accurately comprehend and preserve the semantic content of the input images, with 1 (that the content is entirely non-clothing), 3 (that part of the content is clothing-related), and 5 (that the content is fully clothing-related).

C2: The naturalnes metric evaluates the natural appearance of the images generated by the model. It assesses whether the generated images exhibit any strong sense of incongruity. Images with pronounced unnatural elements are deemed unsuitable for our task of modern adaptive design images for traditional Chinese hanfu. Scores range from 1 (the clothing appears highly unnatural) to 3 (the clothing appears somewhat natural) to 5 (the clothing appears very natural and realistic).

C3: Given our task's objective of generating images for the modern adaptive design of traditional Chinese hanfu, the modern-adaptability of the clothing in the generated images is a key reference indicator. Our goal is to produce garments that are highly suitable for modern lifestyles. The scoring scale ranges from 1 (completely unsuitable for modern daily wear) to 3 (moderately suitable for modern daily wear) to 5 (very suitable for modern daily wear).

C4: Our task emphasizes the preservation and transformation of original hanfu cultural elements. Therefore, cultural-inheritance is another critical reference indicator. Our aim is to generate modernized adaptive clothing images that correctly retain the original hanfu cultural elements. Scores are assigned as follows: 1 (no retention of hanfu cultural elements), 3 (partial retention of hanfu cultural elements), and 5 (retention of most hanfu cultural elements).

C5: As our task is based on clothing design, human subjective aesthetic evaluation is also an assessment indicator. This metric evaluates the final output of the task's pipeline from an aesthetic perspective, helping to determine whether there is room for improvement in terms of aesthetic appeal. The scoring scale ranges from 1 (completely inconsistent with my aesthetic preferences) to 3 (moderately consistent with my aesthetic preferences) to 5 (highly consistent with my aesthetic preferences).

\subsection{Generated Images}\label{generated images}
\begin{figure*}[ht]
    \centering
    \includegraphics[width=1\linewidth]{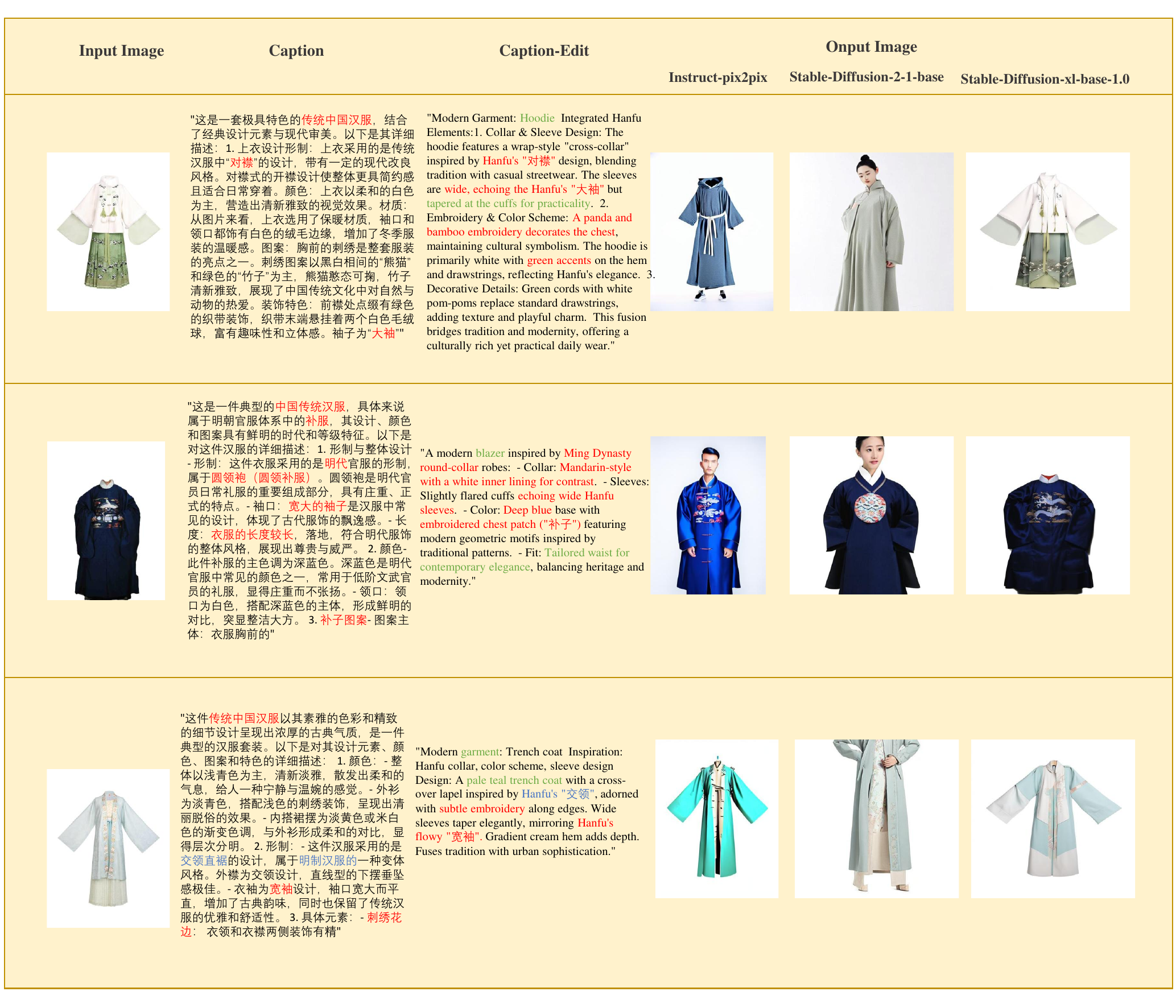}
    \caption{Captions and generated images of cultural image transcreation. The red words in caption are \textcolor{red}{the correct descriptions about the input image} , the blue words in caption are \textcolor{blue}{wrong descriptions}. The green words are \textcolor{green}{the modern adaptions} VLM made in caption-edit, the red in caption-edit are \textcolor{red}{the kept Hanfu cultural elements}, the blue in caption-edit are \textcolor{blue}{the kept Hanfu cultural elements with wrong description}. }
    \label{fig:result-task2}
\end{figure*}

Figure~\ref{fig:result-task2} presents the results of three complete sets of images for the modern adaptive design of the traditional Chinese hanfu process, including the caption recognized and edited by VLM in the middle, and the comparison of images generated by three different models of the same input image, which can intuitively see the image generation effect and preferences of each model.

\subsection{Evaluation Feedback on Generated Images}
Evaluator 1:
Naturalness Perspective: Certain generated images exhibit anomalous features, including human faces synthesized within sleeve areas and garments generated on the heads of figures.
Cultural Perspective: Some outputs misidentify source images as representative of Japanese cultural aesthetics (e.g., anime), resulting in figures and attire resembling classical Japanese 2D manga character designs.

Evaluator 2:
The generated clothing styles demonstrate significant heterogeneity, potentially attributable to outputs produced by distinct generative models. Additionally, the distinction between semantic equivalence and natural authenticity as evaluation criteria remains insufficiently defined.

Evaluator 3:
Most generated garments were identifiable as functional clothing. However, the criterion of "compatibility with contemporary lifestyles" was deemed ambiguous, as interpretations and acceptance thresholds vary substantially across individuals.

Evaluator 4:
Generated outputs exhibited polarization in similarity to source images: either minimal deviation or complete dissimilarity. Furthermore, attempts to integrate traditional cultural elements with modern designs appeared mechanically executed, lacking organic cohesion.

Evaluator 5:
While the majority of generated garments adhered to normative standards, artifacts such as semantically incongruous textual elements or anatomically implausible body parts were observed in certain outputs, detrimentally impacting perceived naturalness.

\label{sec:appendix}

\end{document}